\documentclass[10pt,twocolumn,letterpaper]{article}

\usepackage[pagenumbers]{cvpr} 

\definecolor{cvprblue}{rgb}{0.21,0.49,0.74}
\usepackage[pagebackref,breaklinks,colorlinks,allcolors=cvprblue]{hyperref}
\usepackage{svg}
\usepackage{graphicx}
\usepackage{pifont}

\def\paperID{9816} 
\def\confName{CVPR}
\def\confYear{2026}

\title{CrossJEPA: Cross-Modal Joint-Embedding Predictive Architecture for Efficient 3D Representation Learning from 2D Images}

\author{
Avishka Perera$^{1}$ \quad
Kumal Hewagamage$^{1}$ \quad
Saeedha Nazar$^{1}$ \quad
Kavishka Abeywardana$^{1}$\\
Hasitha Gallella$^{1}$ \quad
Ranga Rodrigo$^{1}$ \quad
Mohamed Afham$^{2}$\\[4pt]
$^{1}$University of Moratuwa \quad
$^{2}$Technische Universität Darmstadt
}

\usepackage{listings}
\usepackage{xcolor}

\lstdefinestyle{mystyle}{
    language=Python,
    basicstyle=\ttfamily\scriptsize,
    keywordstyle=\color{blue},
    stringstyle=\color{green!60!black},
    commentstyle=\color{gray},
    numbers=left,
    numberstyle=\tiny\color{gray},
    stepnumber=1,
    numbersep=10pt,
    backgroundcolor=\color{white},
    showspaces=false,
    showstringspaces=false,
    showtabs=false,
    frame=single,
    breaklines=true,
    tabsize=4,
    captionpos=b
}

\begin{document}
\maketitle
\begin{abstract}

Image-to-point cross-modal learning has emerged to address the scarcity of large-scale 3D datasets in 3D representation learning.
%
However, current methods that leverage 2D data often result in large, slow-to-train models, making them computationally expensive and difficult to deploy in resource-constrained environments.
The architecture design of such models is therefore critical, determining their performance, memory footprint, and compute efficiency.
%
The Joint-embedding Predictive Architecture (JEPA) has gained wide popularity in self-supervised learning for its simplicity and efficiency, but has been under-explored in cross-modal settings, partly due to the misconception that masking is intrinsic to JEPA.
In this light, we propose \textbf{CrossJEPA}, a simple \textbf{Cross}-modal \textbf{J}oint \textbf{E}mbedding \textbf{P}redictive \textbf{A}rchitecture that harnesses the knowledge of an image foundation model and trains a predictor to infer embeddings of specific rendered 2D views from corresponding 3D point clouds, thereby introducing a JEPA-style pretraining strategy beyond masking. 
By conditioning the predictor on cross-domain projection information, CrossJEPA purifies the supervision signal from semantics exclusive to the target domain.
We further exploit the frozen teacher design with a one-time target embedding caching mechanism, yielding amortized efficiency. 
CrossJEPA achieves a new state-of-the-art in linear probing on the synthetic ModelNet40 (\textbf{94.2}\%) and the real-world ScanObjectNN (88.3\%) benchmarks, using only \textbf{14.1M} pretraining parameters (8.5M in the point encoder), and about 6 pretraining hours on a standard single GPU.
These results position CrossJEPA as a performant, memory-efficient, and fast-to-train framework for 3D representation learning via knowledge distillation.
We analyze CrossJEPA intuitively, theoretically, and empirically, and extensively ablate our design choices.
Code will be made available.

\end{abstract}
    
\section{Introduction}
\label{sec:intro}

\begin{figure}[t] 
    \centering
    \includegraphics[width=0.5\textwidth]{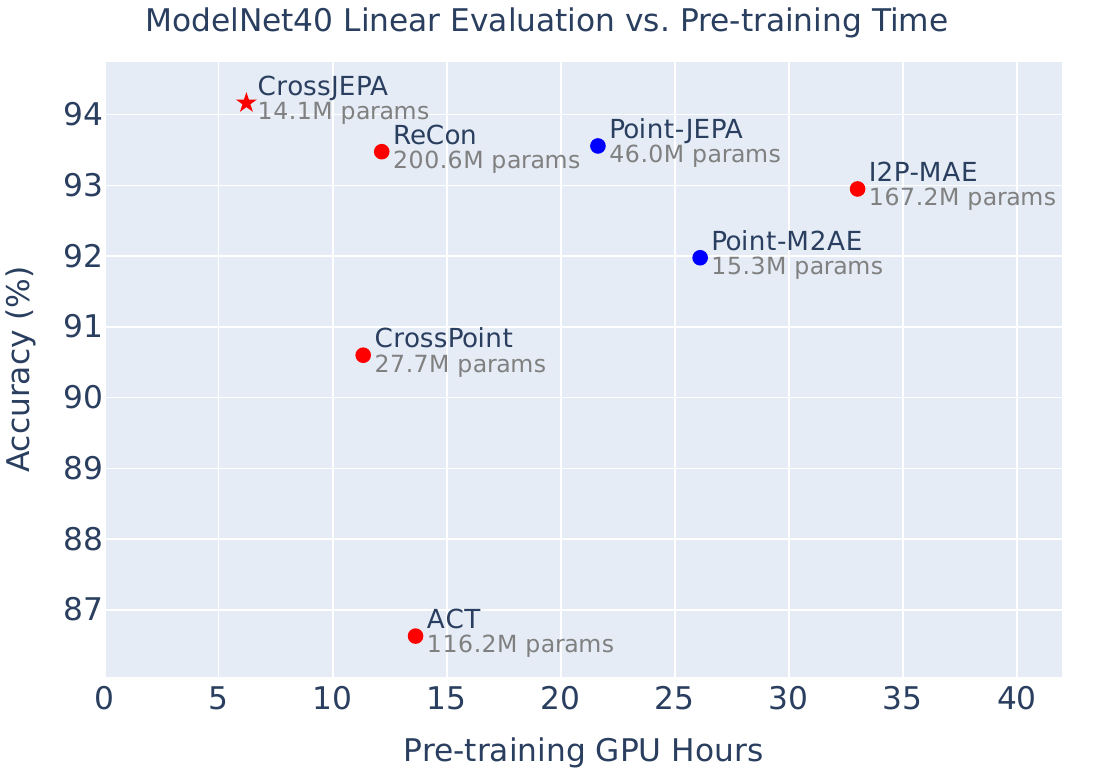}
    \caption{
    \textbf{ModelNet40 Linear Evaluation.}
    Pretraining time on a single NVIDIA RTX 4090 versus best accuracy with an SVM linear classifier on ModelNet40 \cite{modelnet40}. 
    We compare CrossJEPA with recent cross-modal and unimodal point baselines from 2022 onward, covering different SSL architectures: generative \cite{i2p-mae, act, pointm2ae}, joint-embedding architecture (JEA) \cite{crosspoint}, hybrid of generative and JEA \cite{recon}, and JEPA \cite{pointjepa}.
    Red markers denote cross-modal methods, while blue markers denote unimodal point methods.
    CrossJEPA achieves the best trade-off between accuracy, compute, and memory, producing rich point cloud representations with significantly less pretraining compute and fewer parameters.
    Except for CrossPoint \cite{crosspoint}, all other methods use a standard transformer architecture.
    The time required to reach the best accuracy, along with the total number of learnable parameters, is reported (see Table~\ref{table:linear_probing}).
    }
    \label{fig:teaser}
\end{figure}

In 3D vision, effective representation learning from 3D point clouds is essential for real-world applications, including autonomous driving, robotics, mixed reality, and medical imaging \cite{PointCloudSurvey}. 
Given operational constraints in such applications (\eg, deployment on lightweight edge devices), accuracy, memory footprint, and compute efficiency are crucial when developing 3D understanding methods. 
Over the past few years, self-supervised learning (SSL) \cite{PointContrast, pointm2ae, point2vec, pointgpt, pointjepa, locate3d} has significantly advanced performance and efficiency in this domain.

SSL on point clouds has been explored using different architectures, including joint-embedding architectures (JEA) \cite{Info3D, PointContrast, point2vec}, generative architectures \cite{pointgpt, pointmae, pointm2ae}, and other designs \cite{Orientation, recon}. 
Unimodal settings are challenged by (i) scarce large-scale 3D datasets and (ii) sensitivity to hyperparameters that can lead to model collapse \cite{Info3D, point2vec, pointjepa}. 
The absence of strong 3D foundation models, in contrast to image \cite{dinov2, clip} and language \cite{bert, t5} domains, further reflects this.
Recent attempts in 2D-to-3D cross-modal learning \cite{i2p-mae, act, crosspoint, picpoint, recon} address this gap by transferring knowledge from 2D foundation models pretrained on millions of images to point models, thereby stabilizing pretraining. 
Architecture choice, however, directly affects performance, memory, and speed. 
Generative methods \cite{i2p-mae, act} incur higher compute due to point reconstruction, while JEA methods \cite{crosspoint, picpoint} often consume long pretraining epochs to learn invariances.

In contrast, recent work on the joint-embedding predictive architecture (JEPA) has gained attention in SSL for its simple and efficient framework \cite{ijepa, vjepa, pointjepa, brainjepa}, yet it has only been explored in a unimodal setting in point cloud understanding \cite{pointjepa}, where both the context and target encoders operate on point clouds. 
This is partly due to a common misconception that masking, \ie, hiding parts of the input and predicting their latent features from the visible context, is inherently tied to JEPA, inherited from its first concrete application \cite{ijepa}.
However, as per the original introduction of JEPA \cite{lecun_a_path_towards}, masking is \emph{not} a requirement.
Further, naively extending masked regions across modalities is non-trivial and often induces cross-modal inconsistencies. 
We rectify this misconception by introducing \textbf{CrossJEPA}, a simple, lightweight, and efficient \textbf{cross}-modal \textbf{JEPA} that \emph{does not} rely on masking and brings image-to-point supervision into the JEPA community.

Point clouds and images both depict visual characteristics of objects \cite{i2p-mae}, but in different forms.
Just as a child who has only seen photos of apples (2D) can still recognize an apple in the real world (3D), these modalities share underlying semantics.
Motivated by this connection, we draw on the fundamentals of information theory (Sec.~\ref{sec:theoretical_justification}, \emph{Supplementary}) and predictive coding in the brain (Fig.~\ref{fig:predictive_coding}). 
The goal of our work is to strengthen the self-supervision signal to enable more effective learning.
To this end, our architecture combines a frozen image encoder \cite{dinov2}, acting as an \emph{expert} 2D teacher (unlike unimodal JEPA, where teacher and student co-learn \cite{ijepa, pointjepa}), with a point encoder that serves as a 3D student who learns from scratch.
The crucial component is the predictor, which decouples teacher-specific 2D nuisance factors (\eg pose, color), thereby cleansing the gradient flow and guiding the student toward the shared 2D-3D content (\eg, object semantics, shape) to learn richer 3D representations.
Concretely, we train the predictor to infer the image encoder’s embeddings of specific rendered views of a 3D point cloud object, conditioned on 3D-to-2D projection parameters such as yaw and pitch—a pretraining strategy that \emph{does not} involve masking.
Additionally, we exploit the frozen image encoder design by introducing a caching mechanism (Sec.~\ref{sec:caching}, \emph{Supplementary}) for all view embeddings, which eliminates on-the-fly target computation and yields amortized pretraining efficiency, an optimization not adopted by prior cross-modal work \cite{i2p-mae, recon, picpoint, act}.
Together, this simple cross-modal JEPA setup demonstrates a performant, fast-to-train, parameter-efficient pretraining pipeline that produces state-of-the-art (SOTA) results on downstream tasks.
The main contributions of our paper are as follows:
\begin{itemize}
    \item We propose CrossJEPA, which to the best of our knowledge is the first JEPA-style framework for image-to-point cross-modal learning, unlocking JEPA's potential in 3D and enabling \emph{less noisy} knowledge distillation from strong 2D teachers.  

    \item We introduce a new flavor of JEPA \emph{without} explicit masking, based on a challenging latent view prediction task tailored to image and point modalities, which yields more robust point cloud representations.

      \item We show that CrossJEPA achieves superior performance on standard 3D downstream tasks across synthetic and real-world benchmarks, outperforming recent unimodal and cross-modal point models while using only 14.1M pretraining parameters and the least pretraining GPU hours.
    
    \item We provide extensive ablations of our design choices, including image-to-point cross-modal JEPA variants (Sec.~\ref{sec:crossmodal_architecture_selection}, \emph{Supplementary}).
    Additionally, we analyze CrossJEPA from an information theoretic and predictive coding perspective (Sec.~\ref{sec:theoretical_justification}, Sec.~\ref{sec:predictive_coding}, \emph{Supplementary}). 
\end{itemize}


\section{Related Work}
\label{sec:related_work}


SSL based on Energy-Based Models (EBMs) \cite{EBM} emerged as a powerful representation learning approach by leveraging large-scale unlabeled datasets \cite{simclr, BYOL, moco}.

\noindent\textbf{Joint Embedding Architecture (JEA)}. Standard invariance-based models within the EBM framework can be understood as JEAs that map input data into a semantically meaningful space.
Point2Vec \cite{point2vec} employs a teacher-student model for point cloud patch prediction, while several works leverage JEAs to improve point cloud representations \cite{SSL-3DRecSpace,SSL-3DFewShot,SSL-3DDomainAdaptation,JEA2,JEA3}.
CrossPoint \cite{crosspoint} and Pic\texttt{@}Point \cite{picpoint} extend this paradigm to cross-modal image-point learning. In CrossPoint, both teacher and student are learnable, which can make training prone to collapse, while Pic\texttt{@}Point freezes the image encoder but keeps the projection heads learnable. These methods perform well with DGCNN \cite{DynamicGraphCNN} backbones and rely on contrastive learning with long pretraining schedules.
Unlike JEA-based approaches, CrossJEPA uses a frozen 2D teacher and a lightweight predictor to capture structural relationships directly in embedding space, without contrastive pairs or hand-crafted inductive biases.

\noindent \textbf{Generative Architecture.}
EBM-based generative architectures construct target representations directly in the input space.
In 3D, Point-BERT \cite{PointBert} uses variational autoencoders, while Point-MAE \cite{pointmae} and Point-M2AE \cite{pointm2ae} extend masked autoencoding for hierarchical point cloud reconstruction.
Point-GPT \cite{pointgpt}, inspired by GPT \cite{NLPGPT}, introduces autoregressive learning for 3D data, and recent works further refine 3D representations for downstream tasks \cite{3D-Gen1, 3D-Gen2, 3D-Gen3}.
I2P-MAE \cite{i2p-mae} and ACT \cite{act} are generative, masking-based cross-modal distillation methods with frozen image teachers that reconstruct 3D points or 2D features using heavy decoders.
However, such generative models force an EBM to predict low-level details in the input space that are inherently hard to predict, injecting unnecessary training noise, weakening gradient updates for the encoder, and reducing efficiency.
To compensate, such methods require complex pretraining pipelines to suppress these gradient artifacts \cite{i2p-mae}.
In contrast, CrossJEPA avoids masking and applies the loss in the embedding space: a lightweight predictor maps point features using view parameters to frozen image embeddings of specific views, with cached targets removing redundant forward passes.
This sharpens the task relevance of gradient updates, cleaner supervision, giving a simpler architecture with lower compute.

\noindent \textbf{Other architectures.}
Cross modal learning in the point–image space has been explored broadly \cite{recon, pointclip, pointclipv2, Openview, Mambatron, pointcmt}.
Related work such as Image2Point \cite{image2point} and Learning from 2D \cite{2Dlearning} transfers 2D knowledge to 3D via fine-tuning rather than pretraining 3D models from scratch.
ReCon \cite{recon} is a hybrid architecture combining generative and JEA style objectives, coupling masked point reconstruction with cross modal contrastive alignment using text and image teachers.
PointCMT \cite{pointcmt} is another cross-modal knowledge distillation framework where an image teacher and a point student are trained with feature and classifier enhancement losses, using a cross modal point generator for alignment.
By contrast, CrossJEPA adopts a JEPA style predictive framework that \emph{forgoes masking} and uses \emph{precomputed} image embeddings (an optimization not adopted by many prior cross modal baselines with frozen image encoders \cite{i2p-mae, recon, act, picpoint}, as depicted in Fig.~\ref{fig:xjepa-caching}) from a frozen image encoder, aligning with cognitive learning and predictive coding principles (Fig.~\ref{fig:predictive_coding}) \cite{cortical_response,predictive_coding,brain}.

\noindent \textbf{Joint-Embedding Predictive Architecture (JEPA).}
Unlike generative architectures, JEPA \cite{lecun_a_path_towards} optimizes loss in the embedding space, learning predictive relationships between representations rather than enforcing invariance through hand-crafted augmentations. JEPA has been popularly applied across various unimodal setups, including images \cite{ijepa}, videos \cite{vjepa,Vjepa2}, audio \cite{ajepa}, and 3D modalities such as point clouds \cite{pointjepa, 3djepa} and LiDAR \cite{adljepa}. While TI-JEPA \cite{TIjepa} and PrediCIR \cite{PrediCIR} invoke JEPA in cross-modal contexts (text-image), their pretraining remains unimodal (image-only) and relies on latent mask prediction. Recent studies have also investigated multimodal approaches that integrate JEPA for enhanced representation learning \cite{m3jepa}; however, this framework has not been tested for point cloud modality and is not purely cross-modal, as it disentangles information into shared and modality-specific components across multiple modalities. Existing image-point JEPA models are constrained by their single-modality design, being either point-based \cite{pointjepa} or image-based \cite{ijepa}, thereby limiting their ability to leverage complementary structural and contextual information across spatial domains. CrossJEPA fills this gap with an image-point cloud JEPA that introduces a latent view prediction task. This task enables the learning of more robust and semantically rich representations while avoiding model collapse \cite{ijepa} by freezing the image model, offering a new path for advancing SSL in 3D vision.

\section{Methodology}
\label{sec:methodology}

\begin{figure*}[t] 
    \centering
    \includegraphics[width=1\textwidth]{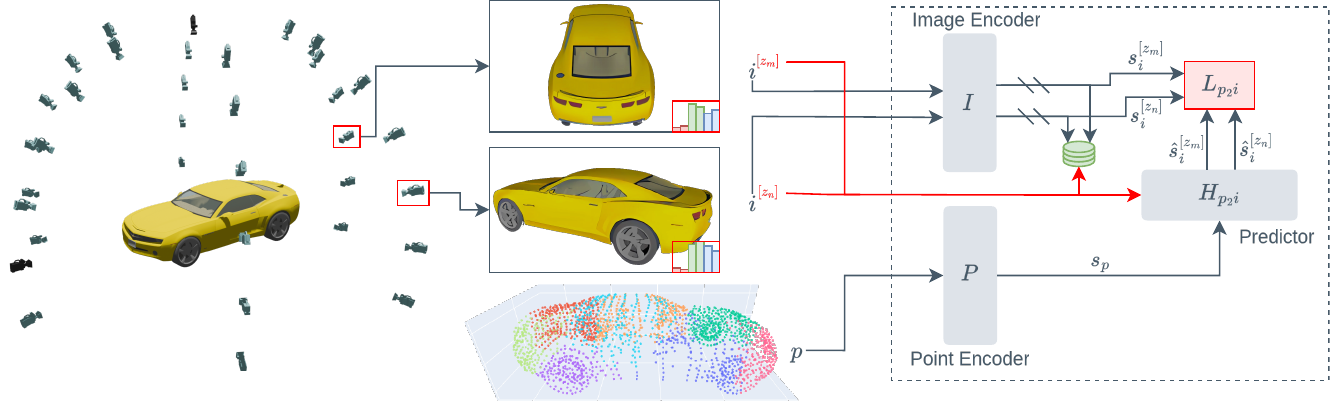} 
    \caption{\textbf{CrossJEPA}:
    CrossJEPA’s training objective is to learn generic features by predicting an image representation $s_i$ from a point cloud representation $s_p$. This is achieved using a distillation architecture that leverages the physical relationship between a point cloud $p$ and its 2D render $i$. The student's learnable point encoder $P$ generates $s_p$, while a frozen, pretrained image encoder $I$ serves as the teacher, providing the target $s_i$. A predictor network is trained to map $s_p$ to specific $s_i^{[z]}$. The image semantics that are unavailable in the point representation are explicitly provided as latent information $z$ to the predictor to make the predictions more definitive.
    Note that the image encoder is frozen; hence, we employ precomputed image embeddings and color histograms as depicted Fig. in~\ref{fig:xjepa-caching} in the \emph{Supplementary}.
    }
    \label{fig:methodology}
\end{figure*}

This section outlines our methodology for training the point encoder by leveraging a pretrained image encoder (\cite{dinov2}) as a supervisor. Our approach employs a predictor network that learns to map point representations to their corresponding image representations. To refine this process, the predictor is explicitly conditioned on latent information unique to the image modality. This conditioning creates a more well-defined objective, significantly reducing noise in the supervision signal propagated back to the point encoder. The specifics of this cross-modal learning strategy are detailed in Section \ref{sec:cross-mo-learning}, while the datasets and implementation details are covered in Section \ref{sec:ds-and-pretrain}. Finally, we discuss the model architecture under \ref{sec:model-arch}.

\subsection{Cross-modal Learning}
\label{sec:cross-mo-learning}

Using a cross-modal setting as our basic supervision architecture, we show that knowledge learnt from a highly generic image model \cite{dinov2} can be effectively transferred to point models. Unlike the typical JEPA predictor (\cite{ijepa, pointjepa, vjepa}) or other predictive approaches (\cite{mae, pointmae, pointm2ae}), we do not mask any part of the input. We further analyze this empirically under the section \ref{sec:masking_ratio}. Instead, we encode the views of a 3D object under 2 degrees of freedom: the pitch and the yaw of the camera. The other parameters of the camera are changed naturally to deliver a visually deterministic placement of the object within the frame. The total number of view renders $|V|$ depends on the number of combinations of yaw ($\theta$) and pitch ($\phi$). We compute 36 ($V$) such views ($i^{[z]}$) per object beforehand. Here, $i^{[z]}$ is the rendered image with a view from a camera placed with a yaw $\theta$ and pitch $\phi$. A single point cloud $p$ is sampled from each object.

We obtain the $s_p$ representation of the point cloud of an object using the $P$ point encoder that we expect to learn. We randomly choose $L$ number of views from the corresponding $36$ total renders ($i^{[z_m]}$, $i^{[z_n]}$, \dots) of the object. Representations of these $L$ views ($s_i^{[z_m]}$, $s_i^{[z_n]}$, \dots) are then obtained from a pre-trained image encoder  $I$. We then attempt to predict these representations by querying the $s_p$ representation by means of a Point-to-Image predictor $H_{p_{2}i}$. This is represented in the Eq. \ref{eq:forward-pass}

This prediction $\hat{s}_i^{[z]}$ is then constrained to \emph{match} with the corresponding target representation. 
Unlike other Image-Point cross-modal methods, which attempt to \emph{align} multi-modal features in a \emph{shared latent space} (\cite{crosspoint, picpoint}), by means of $H_{p2i}$ and the conditioning latent information $z$, we unlock the ability to directly \emph{match} predictions into the target feature space. We analyze the effect of the degree of this information under the Section \ref{sec:latent-information}. Following \cite{ijepa} and \cite{pointjepa}, we use the smooth L1 loss (\cite{smooth_l1} as our learning objective ($L_{p2i}$). For a single object, we apply this criterion for each $v$ image over the manifold $A \subseteq V$. The criterion is then applied for each $k$ sample over the data manifold of all $N$ objects. This is represented in the Eq. \ref{eq:sum-obj}.

\begin{equation} \label{eq:forward-pass}
\begin{split}
    s_p &= P(p) \\
    s_i^{[z_v]} &= I(i^{[z_v]}) \ \text{for } v \in A \subseteq V \\
    \hat{s}_i^{[z_v]} &= H_{p_2i}(s_p, z_v) \ \text{for } v \in A \subseteq V
\end{split}
\end{equation}

\begin{equation} \label{eq:sum-obj}
L_{p_2i} = \sum_{k \in N} \sum_{v \in A \subseteq V} L(\hat{s}_i^{[z_v]}, s_i^{[z_v]})
\end{equation}

The camera pose is encoded using a sinusoidal encoding scheme. The encoding of the yaw and pitch of the target view is similar to the encoding of the horizontal and vertical position of the target patch in I-JEPA \cite{ijepa}. To represent the color histogram, we use 16 bins for each color channel, resulting in a 48-dimensional vector. The camera pose representation is reduced by 48 dimensions to accommodate this vector. 

By conditioning on the pose information at the predictor, we effectively decouple the point encoder from learning the image-specific semantics, which is irrelevant from the point cloud perspective. In other words, gradient components associated with pose or color information act as noise for the point encoder. The introduction of the latent information provides a \emph{sink} for these unnecessary gradient components. As a result, our supervision signal becomes more specific, allowing the encoder to focus exclusively on mutual features. Noisy gradients dilute the learning process and ultimately reduce performance \cite{achille2018emergence, achille2018information, tran2017drgan, katzir2022vector, almudevar2025ib}. This has been empirically proven and has been added to ablation studies under Sec. \ref{sec:latent-information}.

\subsection{Datasets and Pre-processing Setup}
\label{sec:ds-and-pretrain}

To pretrain CrossJEPA, we use ShapeNet \cite{shapenet} (~51.3k CAD models across 55 categories) and Objaverse-XL \cite{objaverseXL} (a large-scale internet 3D dataset with $>$10M assets). We render point clouds and images for 160k Objaverse-XL samples and train jointly with ShapeNet. ShapeNet images are produced via our custom multi-view rendering pipeline.
First, we compute a camera pose matrix for each view using a transformation approach. Given yaw \((\theta)\), pitch \((\phi)\), and a fixed roll \((\rho = 0)\), we generated a rotation matrix \( R \) as:
\begin{equation}
R = R_y(\theta) R_x(\phi) R_z(\rho)
\end{equation}

where \( R_x, R_y, R_z \) are rotation matrices around the X, Y, and Z axes, respectively. The camera position was set at a fixed distance equal to the scene radius \( r \), measured from the scene center \( c \), computed as:

\begin{equation}
P = R \cdot 
\begin{bmatrix} 
0 \\ 
0 \\ 
r \\ 
1 
\end{bmatrix} + c
\end{equation}

The scene radius \( r \) was determined from the bounding box of the object as:

\begin{equation}
r = \| \max(s) - \min(s) \|
\end{equation}

Here, \( s \) represents the set of object vertices.

This method allows for fixing the degrees of freedom of the other camera's extrinsic matrix parameter while varying yaw and pitch. We vary yaw and pitch in \(\pm 60^\circ\) increments, generating 36 different perspectives.
Point clouds are normalized, and scaling, translation, and jittering augmentations are applied. Also, a controlled rotation is applied while preserving the camera pose information embedded in rendering.

\subsection{Model Architecture}
\label{sec:model-arch}
We adopt the patchification strategy from \cite{PointBert}. Initially, the point cloud object is partitioned into $K$ local clusters by applying the k-nearest-neighbor algorithm to a set of pre-selected sub-cloud centers. These clusters are then passed through the point tokenizer (\cite{PointBert}) to convert each cluster into an embedding. A learnable [CLS] token is prepended, forming a token sequence. This representation allows the point cloud to be processed as a sequence by a standard transformer encoder. The final point cloud feature embedding, $s_p$, is obtained by normalizing the embeddings and stacking the max-pooled embeddings with the [CLS] embedding.

We use an input size of 2048 points, yielding 64 clusters, each with 32 points. Our base model uses an embedding dimension of 192 with 18 standard transformer layers. This results in 8.5M parameters for our $P$ point encoder model. Our predictor is also a standard transformer model with an embedding dimension of 192 with 12 layers, finally yielding a total learnable parameter count of 14.1M. We use a frozen ViT-B/14 with DinoV2 \cite{dinov2} weights as our target encoder. For the pre-training setup, we used Adam \cite{Adam} optimizer with a base learning rate of 1.0e-3 following a ramp-up and cosine decay schedule. A 1.0e-6 weight decay was used.
\section{Experiments}
\label{sec:exp}

Following recent work \cite{PointBert, pointmae, pointm2ae, point2vec, pointgpt, pointjepa, crossmost}, to evaluate the effectiveness of CrossJEPA's SSL capabilities, we test the learned point model (\ie, $P$ from CrossJEPA's P2I architecture) against recent top-performing SSL algorithms on four different downstream tasks on standard benchmarks \cite{modelnet40, scanobjectnn}. ModelNet40 \cite{modelnet40} consists of 12311 synthetic 3D objects from 40 distinct categories, while ScanObjectNN \cite{scanobjectnn} contains objects from 15 classes, each containing 2902 unique instances collected by scanning real-world objects. For both datasets, we sample 2048 points per object and sample 64 center points with 32 points in each point patch.
\\

\noindent\textbf{Linear probing}. After pre-training the point encoder, we evaluated it on ModelNet40 \cite{modelnet40} via linear probing. The frozen point encoder's output tokens were pooled using mean and max operations, concatenated with the \texttt{[CLS]} token's output (\texttt{cls+mean+max}), and passed to an SVM classifier. We establish a new \emph{state-of-the-art baseline} for ModelNet40 linear evaluation, with a minimal learnable parameter count. To ensure a fair and direct comparison of pretraining efficiency, we reran several previous works \cite{i2p-mae, pointm2ae, crosspoint, act, recon, pointjepa} alongside CrossJEPA on a single Nvidia RTX 4090. The time to reach the best accuracy was recorded for each experiment. CrossJEPA$_1$ depicts the experiment with 6 parallel view predictions, and CrossJEPA$_2$ depicts the cylindrical pose encoding method. (Table \ref{table:linear_probing})

\begin{table}[h]
\centering
\caption{\textbf{Linear probing on ModelNet40}. Comparison of CrossJEPA with existing SSL methods. \#Params denotes pretraining parameters. Acc is the peak accuracy. GPU T is the time to reach peak accuracy on an NVIDIA RTX 4090. We reproduced results where GPU T is shown; others are cited from the original authors. Cross-modal methods are marked with an asterisk ($^\ast$). CrossJEPA achieves a new state-of-the-art (94.2\%) with only 14.1M parameters. Top metrics are mentioned in \textbf{bold} and the second top is denoted in \textit{italics}.}
\resizebox{\linewidth}{!}{
\begin{tabular}{lccc}
\toprule
\textbf{Method} & \textbf{\#Params} & \textbf{Acc (\%)}& \textbf{GPU T} \\
\midrule
CrossPoint$^\ast$ ~\cite{crosspoint} & 27.7M & 90.6 & 11:20:06 \\
Point-MAE~\cite{pointmae} & 30M & 90.0 & -- \\
Point-M2AE~\cite{i2p-mae} & 15.3M & 92.0 & 26:06:41 \\
I2P-MAE$^\ast$ ~\cite{i2p-mae} & 167.2M & 92.9 & 33:00:50 \\
ACT$^\ast$ \cite{act} & 116.2M & 86.6 & 13:37:55 \\
ReCon$^\ast$ \cite{recon} & 200.6M & 93.5 & 12:08:55 \\
Point-JEPA \cite{pointjepa} & 46.0M & 93.6 & 21:37:46 \\
\midrule
CrossJEPA$_1$$^\ast$ (Ours)& \textbf{14.1M} & \textbf{94.2} & \textit{06:12:51} \\
CrossJEPA$_2$$^\ast$ (Ours)& \textbf{14.1M} & \textit{93.9} & \textbf{03:53:14} \\

\bottomrule
\label{table:linear_probing}
\end{tabular}
}
\end{table}

\begin{table}[ht]
\centering
\caption{\textbf{Linear probing on ScanObjectNN.}  \textbf{\#Points} denotes number of points in the input. \textbf{Acc} is the peak accuracy. Accuracies are cited from original authors. Cross-modal methods are marked with an asterisk ($^\ast$). Top metrics are mentioned in \textbf{bold} and the second top is denoted in \textit{italics}. CrossJEPA achieves \emph{SOTA performance} of 88.3\% accuracy on OBJ-BG setting, despite using fewer parameters, validating the transferability of its learned representations to real-world data.}
\resizebox{0.75\linewidth}{!}{
\begin{tabular}{lcc}
\toprule
\textbf{Method} & \textbf{\#Points} & \textbf{OBJ-BG}  \\
\midrule
STRL  \cite{strl}        & 2k & 77.9   \\
OcCo\cite{occo}          & 2k & 78.3   \\
CrossPoint $^\ast$ \cite{crosspoint}    & 2k & 81.7   \\
Pic@Point \cite{picpoint}     & 2k & 85.7   \\
ACT $^\ast$  \cite{act}         & 2k & 85.2 \\
ReCon $^\ast$ \cite{recon}       & 2k & \textbf{89.5}  \\
Point-JEPA \cite{pointjepa} &-- &-- \\
DS-MAE \cite{dsmae}       & 2k & 86.5   \\
I2P-MAE$^\ast$ \cite{i2p-mae}      & 2k & 87.1   \\
\midrule
\textbf{CrossJEPA$^\ast$(Ours)}        & 2k & \textit{88.30}  \\
\bottomrule
\end{tabular}
}
\end{table}

\noindent\textbf{Fine-tuning}. We also investigate the performance of the learned representation via end-to-end fine-tuning. After pre-training, we utilize the point encoder ($P$) to extract the max and average pooled outputs along with the cls token output. These outputs are then processed by a three-layer MLP for classification tasks. This class-specific head, as well as the point encoder, is fine-tuned end-to-end on ModelNet40.

\begin{table}[h]
\centering
\caption{\textbf{Linear classification with fine-tuned weights on ModelNet40~\cite{modelnet40}}. \textbf{\#Points} denotes the number of points in the input. \textbf{Acc} is the peak accuracy. Accuracies are cited from the original authors. Cross-modal methods are marked with an asterisk ($^\ast$). CrossJEPA performs on par (94.3\%)  with existing SOTA. Top metrics are mentioned in \textbf{bold} and the second top is denoted in \textit{italics}.}
\begin{tabular}{lcc}
\toprule
\textbf{Method} & \textbf{\#Points} & \textbf{Acc (\%)} \\
\midrule
Transformer + OcCo~\cite{occo} & 1k & 92.1 \\
Point-BERT \cite{PointBert} & 8k & 93.8 \\
Point-M2AE~\cite{pointm2ae} & 1k & 94.0 \\
I2P-MAE$^\ast$ ~\cite{i2p-mae} & 1k & 93.7  \\
ACT$^\ast$ \cite{act} & 1k & 93.7  \\
ReCon$^\ast$ \cite{recon} & 1k & 94.1  \\
ReCon$^\ast$ \cite{recon} & 8k & \textbf{94.3}  \\
Point-JEPA \cite{pointjepa} & 1k & 93.8 \\
\midrule
CrossJEPA$^\ast$ (Ours) & 2k & \textbf{94.3} \\
\bottomrule
\label{table:model_net_finetune}
\end{tabular}
\end{table}

\noindent\textbf{Few-Shot Learning.}
We conducted few-shot learning experiments on ModelNet40 following the m-way--n-shot protocol, evaluating 5-way and 10-way settings with 10-shot and 20-shot configurations. Our model achieved state-of-the-art performance across all settings, notably outperforming several SOTA methods (Table \ref{table:fewshot}).

\begin{table}[h]
    \centering
        \caption{\textbf{Few-shot classification on ModelNet4.} We report 5-way and 10-way tasks under 10-shot and 20-shot protocols. Cross-modal methods are marked with an asterisk ($^\ast$).  Top metrics are mentioned in \textbf{bold} and the second top is denoted in \textit{italics}. CrossJEPA dominates all prior work by a large margin, achieving state-of-the-art few-shot performance.}
    \renewcommand{\arraystretch}{1.2}
    \resizebox{\columnwidth}{!}{
    \begin{tabular}{p{3.2cm}cccc} 
        \toprule
        \textbf{Method} & \multicolumn{2}{c}{\textbf{5-way}} & \multicolumn{2}{c}{\textbf{10-way}} \\
        \cmidrule(lr){2-3} \cmidrule(lr){4-5}
        & \textbf{10-shot} & \textbf{20-shot} & \textbf{10-shot} & \textbf{20-shot} \\
        \midrule
        Point-BERT\cite{PointBert} & 94.6\,$\pm$\,3.1 & 96.3\,$\pm$\,2.7 & 91.0\,$\pm$\,5.4 & 92.7\,$\pm$\,5.1 \\
        CrossPoint $^\ast$\cite{crosspoint} & 92.5\,$\pm$\,3.0 & 94.9\,$\pm$\,2.1 & 83.6\,$\pm$\,5.3 & 87.9\,$\pm$\,4.2 \\
        Point-MAE\cite{pointmae} & 96.3\,$\pm$\,2.5 & 97.8\,$\pm$\,1.8 & 92.6\,$\pm$\,4.1 & 95.0\,$\pm$\,3.0 \\
        Point-M2AE\cite{pointm2ae} & 96.8\,$\pm$\,1.8 & 98.3\,$\pm$\,1.4 & 92.3\,$\pm$\,4.5 & 95.0\,$\pm$\,3.0 \\
        ReCon $^\ast$\cite{recon} & 97.3\,$\pm$\,1.9 & 98.9\,$\pm$\,1.2 & 93.3\,$\pm$\,3.9 & 95.8\,$\pm$\,3.0 \\
        ACT$^\ast$\cite{act}    & 96.8\,$\pm$\,2.3 & 98.0\,$\pm$\,1.4 & 93.3\,$\pm$\,4.0 & 95.6\,$\pm$\,2.8 \\
        Point2Vec \cite{point2vec}& 97.0$\pm$2.8 & 98.7$\pm$1.2 & 93.9$\pm$4.1 & 95.8$\pm$3.1 \\
        Point-JEPA \cite{pointjepa} & \textit{97.4$\pm$2.2} & \textbf{99.2$\pm$0.8} & \textbf{95.0$\pm$3.6} & \textit{96.4$\pm$2.7}\\
        \midrule
        \textbf{CrossJEPA}$^\ast$ (Ours)& \textbf{97.8$\pm$2.5} & \textit{99.0$\pm$1.0} & \textit{95.0$\pm$4.4} & \textbf{96.5$\pm$3.1} \\

        \bottomrule
    \end{tabular}
    }

    \label{table:fewshot}
\end{table}

\noindent\textbf{Part Segmentation.}
We evaluate CrossJEPA on the ShapeNetPart \cite{shapenet} segmentation task, following prior work \cite{pointmae, PointBert, pointm2ae, pointjepa} and using the Point2Vec \cite{point2vec} segmentation head. This head combines global and per-point features. We source these by averaging embeddings from our encoder's 6th, 12th, and 18th blocks. The global vector is formed by pooling this average and appending the [CLS] token and class label. The per-point features are formed by up-sampling this average using PointNet++ \cite{pointnet++}. The global vector is then concatenated to each per-point vector and passed to a shared MLP for prediction. While Tab. \ref{table:segmentation} shows competitive results, CrossJEPA's performance is lower than SOTA, a result we attribute to our smaller encoder size and use of fewer point clusters.

 \begin{table}[h]
\centering
\caption{\textbf{ShapeNetPart segmentation performance.} Results are reported for unseen categories (mIoU\(_C\)) and unseen instances (mIoU\(_I\)). \textit{Cross-modal methods} are marked with an asterisk. CrossJEPA achieves competitive segmentation scores.}
    \begin{tabular}{lcc}
        \toprule
        \textbf{Method} & \textbf{mIoU\textsubscript{C}} & \textbf{mIoU\textsubscript{I}} \\
        \midrule
        Transformer-OcCo \cite{occo} & 83.4 & 85.1 \\
        Point-BERT \cite{PointBert} & 84.1 & 85.6 \\
        Point-MAE \cite{pointmae} & 84.1 & 86.1 \\
        CrossPoint$^\ast$ \cite{crosspoint} & - & 85.5\\
        Point-M2AE \cite{pointm2ae} & \textit{84.9} & \textit{86.5} \\
        Point2Vec \cite{point2vec} & 84.6 & 86.3 \\
        PointGPT-S \cite{pointgpt} & 84.1 & 86.2 \\
        I2P-MAE $^\ast$  & \textbf{85.15} & \textbf{86.76}\\
        Point-JEPA  & 83.9 & 85.8\\
        \bottomrule
        CrossJEPA$^\ast$ (Ours) & 
        84.1 & 85.8 \\
        \bottomrule
    \end{tabular}
    \label{table:segmentation}
\end{table}

\section{Ablations}
\label{sec:ablations}

We evaluate the ModelNet40 classification accuracy by changing variables in our pretraining setup. We validate our point model after each training loop. In each table, \textbf{Acc.} represents the percentage of ModelNet40 classification accuracy, \textbf{Sat. Epoch} denotes the epoch on which the maximum accuracy was obtained, \textbf{Epoch T.} denotes the average time per epoch, \textbf{\#Params} denotes the total number of parameters during training.
\\

\begin{figure}[h]
    \centering
        \caption{\textbf{Decay of the learning loss at different latent information configurations.} This diagram depicts how the discrepancy between the image representation (target) and the model prediction reduces with epochs. By increasing the latent information, the training loss reduced drastically.
}
    \includegraphics[width=0.45\textwidth]{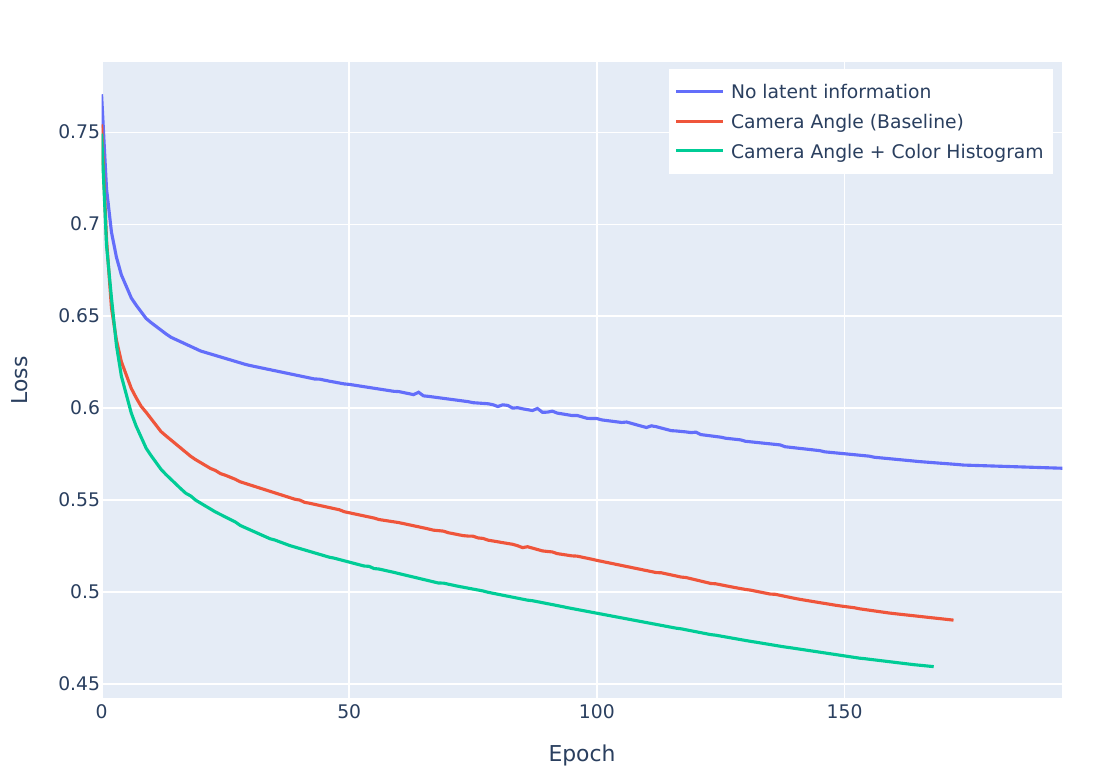}

    \label{fig:latent-info}
\end{figure}

\noindent\textbf{Latent Information Aided Representation Prediction.} \label{sec:latent-information}
Three tests were conducted to evaluate whether additional latent information makes any difference for the pretraining task. We first evaluate the configuration that provides the camera pose of the rendered view to the prediction process. Then, an ablation experiment is performed by removing this pose information. To further evaluate the significance of the latent information, we provide the color histograms of the target image to the prediction process. For the ablation experiment, we replace these sinusoidal camera pose embeddings with a fixed zero vector.
The Table \ref{table:latent-info} and Fig. \ref{fig:latent-info} present the effect of providing \emph{latent information} to the predictor. This \textit{latent information} is necessarily information exclusive to the target domain and unavailable in the source domain. The results can be interpreted both intuitively and from an information theoretic perspective as follows.

\begin{table}
    \centering
        \caption{\textbf{Effect of latent information for predictor in JEPA.} Adding camera pose and color histogram information consistently improves both accuracy and convergence speed. The best configuration yields 94.0\% accuracy with fewer epochs, confirming that CrossJEPA leverages latent cues to accelerate learning and reach world-leading performance.}
    \renewcommand{\arraystretch}{1.3}
    \resizebox{0.9\columnwidth}{!}{
    \begin{tabular}{lcc} 
        \toprule
        \textbf{Experiment} & \textbf{Acc.} & \textbf{Sat. Epoch} \\ 
        \midrule
        No Latent Info & 93.68\% & 93 \\
        Cam Pose & 93.84\% & 71 \\
        Cam Pose + Color Hist.  & 94.04\% & 67 \\
        \bottomrule
    \end{tabular}
    }

    \label{table:latent-info}
\end{table}

\noindent\textbf{1) Intuitively:} By providing the latent information to the predictor, we will be \emph{asking more specific questions} from the predictor, \eg, instead of asking for the ``image representation of a car", we will be asking ``image representation of the front view of a yellow car". This will make the prediction more specific, eventually creating a less noisy supervision signal.

\noindent\textbf{2) Information theory-based:} The image representation is composed of multiple components. Putting them in a generic sense: high-level object-centric semantics and low-level image-specific details. While the object semantics provide a rich supervisory signal for the point cloud model, the image-specific details are irrelevant to the 3D task and act as a source of noise. To address this, we explicitly provide these low-level features as direct inputs to the predictor. This creates a dedicated pathway that serves as a sink for the corresponding gradients, limiting them from propagating backward. As a result, these nuisance signals are decoupled from the primary learning signal, yielding a cleaner and more effective supervision for the point model. This perspective is theoretically justified in the \emph{Supplementary} (Sec. \ref{sec:theoretical_justification}).
The above empirical results suggest that the more latent information we provide, the better the accuracy becomes, and the lower the time it takes for an encoder to understand the underlying semantics. However, theoretically, the degree of this latent information has an upper bound for an optimal learning strategy, even though we fail to observe it with the current evaluation setting. Defining such ``exclusive target-specific information" can be challenging. Even though some information is unavailable in the source domain, the context encoder can attempt to predict that, making the overall training efficient. \Eg, It's better if the point model can predict that the color of a leaf is green rather than blue. But by explicitly providing this \textit{pseudo learnable information}, we might be limiting the method's capabilities. 
\\

\noindent\textbf{Masking ratio.}\label{sec:masking_ratio}
Many contemporary instantiations of JEPA show that the masking percentage has a "sweet spot". The reason for such a phenomena is a trade-off balancing problem. Since those algorithms \cite{ijepa, vjepa, pointjepa} solely rely on the masking strategy for training, a significant amount of masking is required for the supervision to take effect. On the other hand, continuation of this increment reduces the information of the data while creating a discrepancy between the student and teacher, and also between the learning and inference modes, yielding a lower inference performance. 

But CrossJEPA uses an entirely different and new flavour of JEPA, which completely removes the first side of the aforementioned trade-off problem. Making it possible to feed the full input while training. Hence, the encoder and predictor find the correct relationships between the tokens. This argument is clearly reflected by the fact that the results steadily drop with increasing masking ratio in Table \ref{table:masking-ratio}. This also supports our claim that masking may harm cross-modal pretraining due to cross-modal inconsistencies.
\\

\begin{table}
    \centering
        \caption{\textbf{Effect of masking ratio:} Accuracy remains stable across different masking levels, while training becomes more efficient at higher ratios. These results show that CrossJEPA tolerates aggressive masking without degradation, supporting its scalability for large-scale training.}
    \resizebox{0.75\columnwidth}{!}{
    \begin{tabular}{lcc} 
        \toprule
        \textbf{Masking Ratio} & \textbf{Acc.} & \textbf{Epoch T.} \\ 
        \midrule
        0 & 93.96\% & 02:34\\
        0.3 & 93.84\% & 02:20  \\
        0.5  & 93.55\% & 02:13  \\
        0.75  & 93.51\% & 02:02  \\
        \bottomrule
    \end{tabular}
    }

    \label{table:masking-ratio}
\end{table}

\noindent\textbf{Parallel View Prediction}. We ablate predicting multiple image representations during a single iteration in Table \ref{table:parallel-view-prediction-transposed}.
The results illustrate the fundamental bias-variance trade-off.
Note that, despite increasing the simultaneous view count, the training time increases marginally, showing the efficiency of our image embedding caching strategy.
\\

\begin{table}
    \centering
        \caption{\textbf{Effect of parallel view prediction count:} Accuracy, epoch duration, and saturation epochs for different numbers of simultaneous views (1–36). Results show that while accuracy remains stable across view counts, increasing the number of views reduces training efficiency by raising the per-epoch cost. With more views, the model saturates earlier due to redundant supervision signals, highlighting a trade-off between multi-view consistency and computational cost.}
    \resizebox{0.9\columnwidth}{!}{
    \begin{tabular}{lccc}
        \toprule
        \textbf{View Count} & \textbf{Acc.} & \textbf{Epoch T.} & \textbf{Sat. Epoch} \\
        \midrule
        1 & 93.76\% & 160 & 02:18 \\
        6 & 94.17\% & 160 & 02:18 \\
        12 & 94.04\% & 157 & 02:29 \\
        18 & 93.92\% & 110 & 02:39 \\
        24 & 93.84\% & 116 & - \\
        30 & 93.80\% & 122 & 02:56 \\
        36 & 93.71\% & 115 & 02:56 \\
        \bottomrule
    \end{tabular}
    }
    \label{table:parallel-view-prediction-transposed}
\end{table}

\noindent\textbf{Model Scaling}
To analyze the effect of model capacity, we compare three configurations: \textbf{Small}, \textbf{Base}, and \textbf{Large}. 
The results in Table~\ref{tab:model-scaling} show a clear performance gain as the number of parameters increase. 

\begin{table}[ht]
\centering
\caption{\textbf{Effect of model scaling}. Accuracy and model size are reported for Small, Base, and Large variants. Increasing model size improves accuracy, though at the expense of parameter count.}
\label{tab:model-scaling}
\resizebox{\linewidth}{!}{
\begin{tabular}{lcccc}
\toprule
\textbf{Model Size} & \textbf{Acc.} & \textbf{\#Params} & \textbf{Sat. Epoch} & \textbf{Epoch T.} \\
\midrule
Small & 93.72 & 6.4M & 118 & 02:26 \\
Base  & 93.84 & 14.1M & 71	& 02:20 \\
Large & 94.04 & 24.8M  & 51	& 03:33\\
\bottomrule
\end{tabular}
}
\end{table}

%
%
%


\section{Discussion}
\label{sec:discussion}

\textbf{Is CrossJEPA, a JEPA?}
Yes. JEPA’s core is simple: predict target representations from context in representation space; target and context can be of different modalities \cite{lecun_a_path_towards}.
\\

\noindent\textbf{Shouldn't JEPA contain masking?}
No. 
The core JEPA framework is built on four generic criteria from an information-theoretic perspective (see Fig.~13 in \cite{lecun_a_path_towards}) and \emph{does not} require masking. 
I-JEPA's masking strategy is a \emph{particular instantiation} used to validate the paradigm---not the paradigm itself. 
When the architectural need of masking is not present, as in our case, we show that masking can also lead to cross-modal inconsistencies hurting performance, as we report in Sec.~\ref{sec:masking_ratio}.
\\

\noindent\textbf{The architecture is simple.}
JEPA is a simple SSL architecture; thus, the final design of our work is indeed simple, which we view as an advantage. This was achieved after extensive ablations, as detailed in the \emph{Supplementary} (Sec.~\ref{sec:crossmodal_architecture_selection}), where we investigated complex architectural variants before converging on this highly efficient, effective, and focused solution.
\\

\noindent\textbf{Why fix the rendered views?}
Camera parameters are continuous values by definition, but the well-established positional encoding scheme \cite{vaswani2023attentionneed} requires discrete values.
\section{Conclusion}
\label{sec:conclusion}

In this work, we introduce CrossJEPA, a cross-modal JEPA that effectively distills 2D knowledge to learn rich 3D point cloud representations. 
We establish the conceptual basis of our framework, whose deliberate simplicity is validated by achieving SOTA results on diverse 3D tasks with the least GPU pretraining hours and parameter counts, thereby promoting sustainable AI through resource-conscious research approaches.
A current limitation of CrossJEPA is its reliance on fixed projection parameters; thus, future work can explore adaptive or learnable projection strategies.
%
Extending the CrossJEPA to richer multimodal setups, such as combining LiDAR and camera data for autonomous driving \cite{mei2022waymoopendatasetpanoramic, Geiger2013IJRR}, is another promising direction.
{
    \small
    \bibliographystyle{ieeenat_fullname}
    \bibliography{main}
}

\clearpage
\setcounter{section}{0} 
\renewcommand*{\thesection}{\Alph{section}}
 \maketitlesupplementary
 
\begin{figure*}[!t] 
    \centering
    \includegraphics[width=1.1\textwidth]{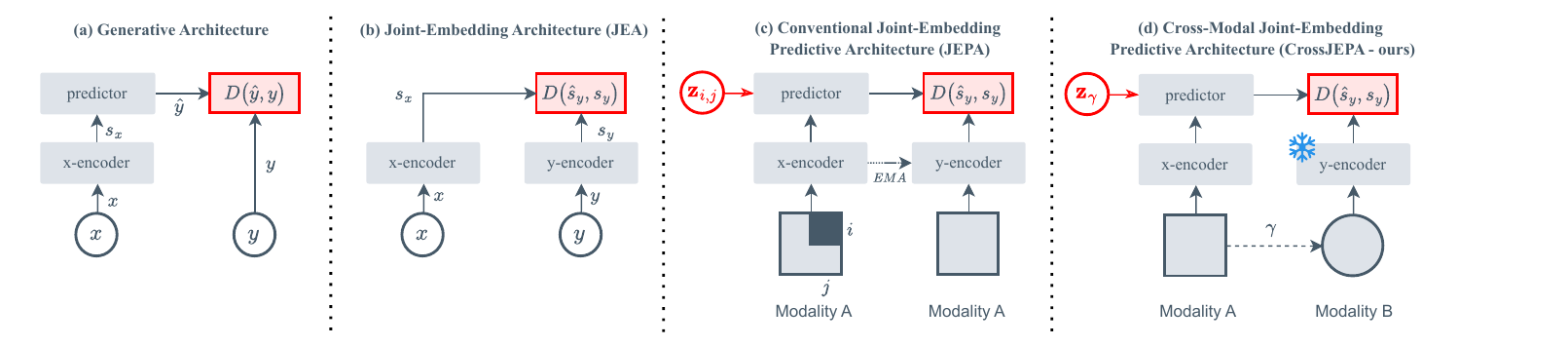}
    \caption{\textbf{Common SSL Architectures}.
    (a) Generative architectures directly predict $\hat{y}$ from the representation $s_x$ of input $x$ (where $x$ is often a corrupted version of $y$). 
    (b) Joint-Embedding Architectures (JEA) minimize the embedding distance between $s_x$ and $s_y$ for signals $x$ and $y$, typically through invariance-based learning. 
    (c) Joint-Embedding Predictive Architectures (JEPA) use a predictor conditioned on minimal latent information $z$ to predict the abstract representation $s_y$ instead of reconstructing $y$, thereby focusing on semantic structure rather than modality-specific details. 
    Conventional JEPA pretraining relies on masking a single modality $A$ by selecting random patches $[i,j]$ and predicting the embeddings of the masked regions, which limits JEPA to unimodal setups since mask correspondence across modalities introduces inconsistencies. 
    (d) \textbf{CrossJEPA} moves beyond masking through a latent view prediction objective. 
    The input modality $A$ (\eg, 3D point cloud) is projected into modality $B$ (\eg, 2D images) using cross-domain projection parameters, and the predictor learns to infer the representation of $y$ in modality $B$ conditioned on latent information $\gamma$ that is exclusively of modality $B$. 
    This yields a cleaner learning signal, removes modality-$B$-specific information originated in the frozen $y-encoder$ (modality $B$ teacher) from the gradient path, and enables the point student to focus on mutual cross-modal information to learn richer 3D representations.}
    \label{fig:architectural_comparison}
\end{figure*}

\section{Cross-modal Architecture Selection}
\label{sec:crossmodal_architecture_selection}

SSL in 2D and 3D has evolved through three major architectural families. Generative architectures (Fig.~\ref{fig:architectural_comparison} (a)) \cite{pointmae,pointm2ae,pointgpt,i2p-mae,act} learn by reconstructing input signals, requiring heavy decoders and forcing models to predict low-level visual details—often introducing noise and inefficiencies. Joint-Embedding Architectures (JEA) (Fig.~\ref{fig:architectural_comparison} (b)) \cite{PointContrast,point2vec,crosspoint,picpoint} instead align representations of augmented views or cross-modal pairs through contrastive or matching objectives, but rely heavily on invariance engineering and remain susceptible to long training schedules or collapse when teachers are learnable. Joint-Embedding Predictive Architectures (JEPA) (Fig.~\ref{fig:architectural_comparison} (c)) \cite{lecun_a_path_towards,ijepa,vjepa,pointjepa} shift the learning signal to predicting abstract latent representations instead of raw data, enabling semantic-focused supervision. However, conventional JEPA implementations are unimodal, typically applied via latent mask prediction on a single modality, which prevents direct extension to image–point cloud settings because mapping mask correspondences across modalities is non-trivial. CrossJEPA (Fig.~\ref{fig:architectural_comparison} (d)), in contrast, introduces a JEPA-style latent view prediction objective that avoids masking entirely and uses a frozen 2D teacher to provide stable, high-level supervision signals. By predicting latent image representations of rendered 3D views, conditioned only on minimal cross-domain projection information, CrossJEPA delivers a cleaner learning signal, removes target modality-specific noise from the gradient path, and enables efficient, robust cross-modal representation learning.

\begin{figure}[h]
    \centering
    \small
    \includegraphics[width=0.5\textwidth]{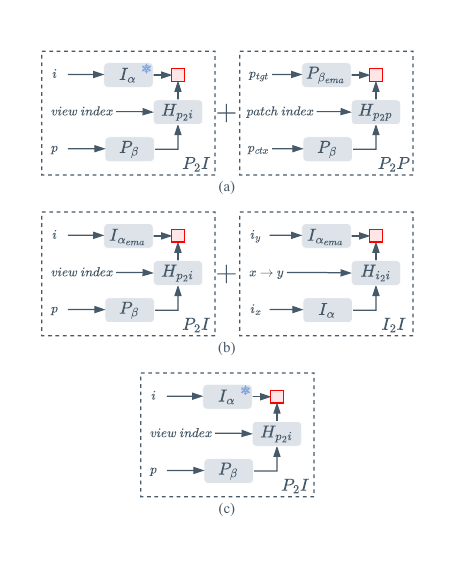}
    \caption{\textbf{Cross-modal Architecture Selection} We explore three different cross-modal architectures: (a) $P_2I$ and $P2P$ learning objectives. (b) $P_2I$ and $I2I$ learning objectives. (c) $P2I$ leaning objective alone. The $P_2I$ learning objective attempts to predict the representations of views of an object specified by a certain view index $\gamma$ by referring to a point cloud representation. $P_2P$ learning objective is the same as \cite{pointjepa}: the representations of masked point groups are predicted by referring to a context point group. $I_2I$ learning objective attempts to predict the representation of a target view ($y$) by referring to a context view ($x$).}
    \label{fig:xjepa-architectures}
\end{figure}

We explored three different cross-modal architectures, as depicted in Fig. \ref{fig:xjepa-architectures}, prior to pursuing our proposed method. They are as follows:
\begin{itemize}
    \item $P_2I + I_2I$ (CrossJEPA with image domain learning)
    \item $P_2I + P_2P$ (CrossJEPA with point and image teachers)
    \item $P_2I$ (Pure CrossJEPA with frozen image encoder)
\end{itemize}

\begin{figure*}[] 
    \centering
    \includegraphics[width=0.77\textwidth]{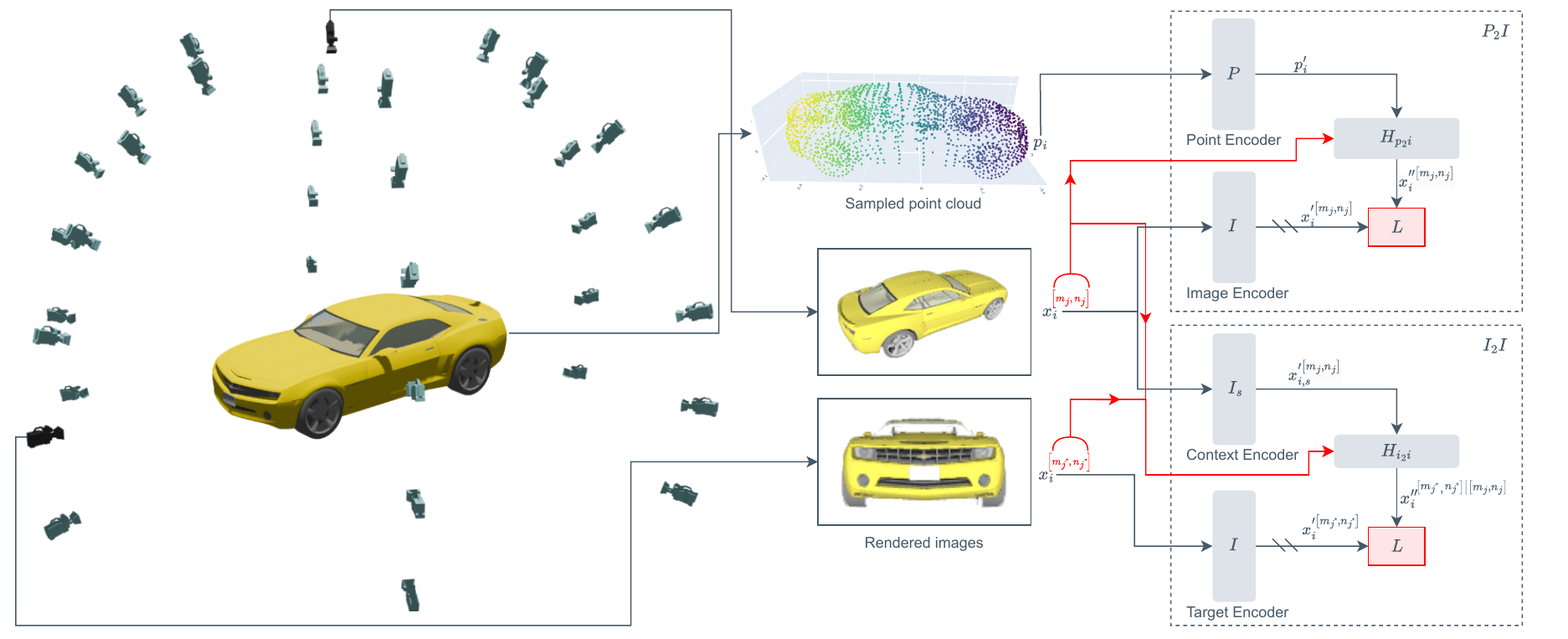} 
    \caption{
        \textbf{Proposed P2I+I2I framework} with two parallel branches: 
        \textbf{Point-to-Image ($P_2I$)} and \textbf{Image-to-Image ($I_2I$)} learning. 
        In $P_2I$, the model predicts an image representation given a point cloud and a camera view (yaw, pitch), and optimizes the point encoder using a smooth L1 loss. 
        In $I_2I$, a student image encoder predicts the representation of a target view based on a source view and corresponding camera parameters. 
        The student encoder is trained via the same loss, while the teacher is updated using EMA.
    }
    \label{fig:i2i}
\end{figure*}

Next, we will explore the architectural setups that we omitted.

\subsection{P2I + I2I (CrossJEPA with Image Domain Learning)}
\label{sec:p2i_i2i}

\begin{figure*}[] 
    \centering
    \includegraphics[width=\textwidth]{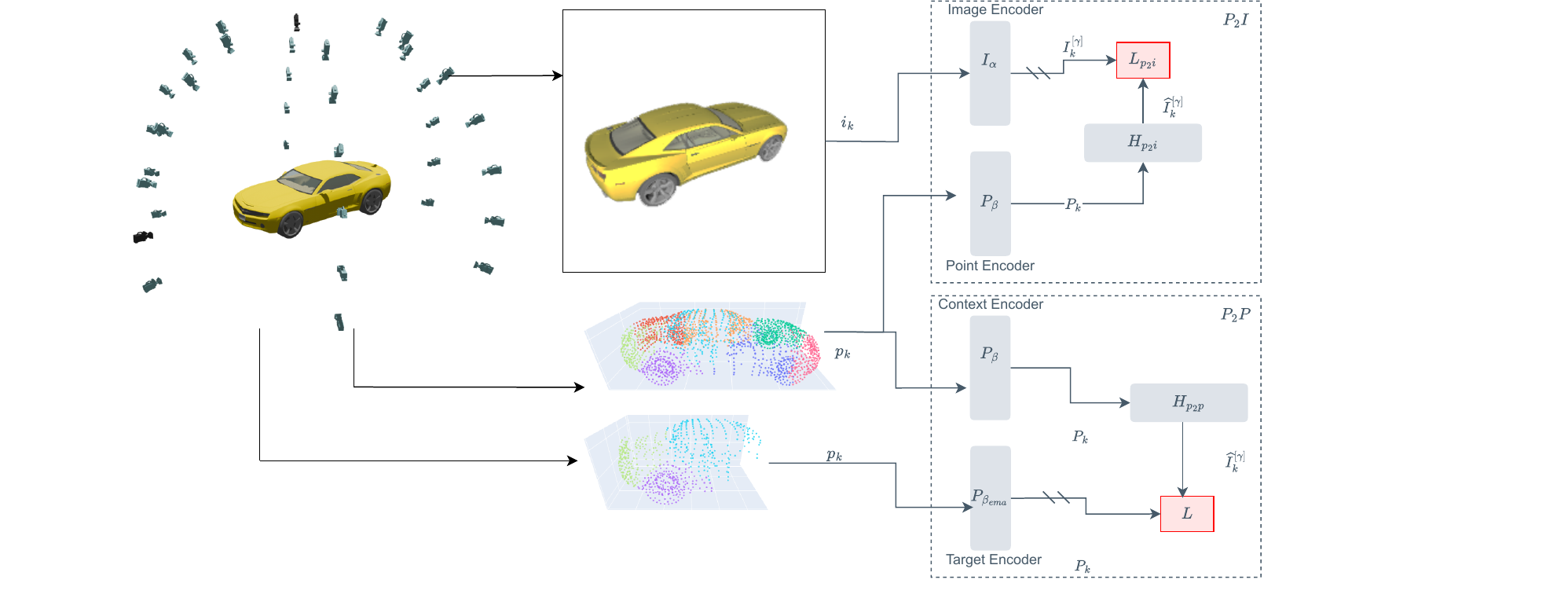} 
    \caption{
        \textbf{Proposed P2I+P2P framework} with two parallel branches: \textbf{Point-to-Image (P$_{2}$I)} and \textbf{Point-to-Point (P$_{2}$P)} learning. 
    In P$_{2}$I, the model predicts an image representation from a point cloud and a view token (yaw, pitch), supervised using features from a frozen image encoder. 
    The predictor is trained using a smooth L1 loss. 
    In P$_{2}$P, the model learns view-consistent point cloud representations by predicting the target view’s embedding from a source point cloud using a relative pose token. 
    The student point encoder is trained with the same loss, while the target encoder is updated using EMA. 
    Together, these branches enable cross-view supervision across both image and point modalities.
    }
    \label{fig:p2p}
\end{figure*}

While we start off with a pre-trained image model \cite{dinov2}, we utilize an online fine-tuning setup that was inspired by \cite{ijepa} and \cite{crossmost}. These methods follow the masked or augmented patch prediction method. However, we do not use any sort of augmentations in the input space. We rather exploit the fact that we have multiple known views for each object within our training set. Formally, we have an $\mathbf{I}_{\alpha}$ learnable image model, and an $\mathbf{I}_{\alpha_{ema}}$ model whose weights are updated as an exponential moving average of the $\mathbf{I}_{\alpha}$ model with a momentum of $\lambda$. See Fig.~\ref{fig:i2i}.

\begin{equation} \label{eq:ema}
    \begin{split}
        \mathbf{I}_{\alpha_{ema}} = \mathbf{I}_{\alpha_{ema}} \times\lambda+\mathbf{I}_{\alpha}\times(1-\lambda)
    \end{split}
\end{equation}

For a sequence of randomly sampled $k$ views of the same object, we obtain the corresponding representations of both student and teacher models. Using the representations of the student model ($I_k^{\alpha}$) we attempt to predict the representations of the teacher model by aligning the predictions ($\hat{I}_k$) with the teacher model's output ($I_k^{\alpha_{ema}}$) using the smooth loss ($L$). We obtain this prediction by querying an image predictor model ($\mathbf{H}_{i_2i}$) using the relative view indexes as latent variables ($\boldsymbol{\gamma}^{\alpha} \rightarrow \boldsymbol{\gamma}^{\alpha_{ema}}$). See Figure \ref{fig:i2i}

\begin{equation}
\begin{split}
    I_k^{\alpha} &= \mathbf{I}_{\alpha}(i_k) \\
    I_k^{\alpha_{ema}} &= \mathbf{I}_{\alpha_{ema}}(i_k) \\
    \hat{I}_k &= \mathbf{I}_{i_2i}(I_k^{\alpha}, \boldsymbol{\gamma}^{\alpha}\rightarrow \boldsymbol{\gamma}^{\alpha_{ema}}) \\
    \mathcal{L}_{i_2i} &= \sum_k \sum_{s, t} \mathcal{L}(I_k^{\alpha_{ema}}, \hat{I}_k)
\end{split}
\end{equation}

\noindent
Here, $\boldsymbol{\gamma}^{\alpha}$ and $\boldsymbol{\gamma}^{\alpha_{ema}}$ represent the view indexes corresponding to the student and teacher model inputs, respectively.

\noindent
$s$ and $t$ index over all student and teacher model outputs, respectively.

We follow a cosine schedule for the EMA by starting from 1 and then complete a full cycle with a minimum value of 0.996. In order to retain the knowledge of the image model amidst the noisy initialization of the predictor, we start off from a high momentum.

\begin{equation}
    \mathcal{L} =\mathcal{L}_{p_2i}+\mathcal{L}_{i_2i}
\end{equation}

According to the above equation, we finally produce the final learning objective as minimizing the sum of $L_{p_2i}$ and $L_{i_2i}$. However, this architectural setup demonstrated inferior performance due to potential model collapse in the learnable image model component.

\subsection{P2I + P2P (CrossJEPA with Point and Image Teachers)}
\label{sec:p2i_p2p}

Unlike the previous setting, where both image branches are optimized jointly, here we freeze the pretrained image model and supervise the learning solely through the point domain. The model is trained using two parallel paths: a \textit{point-to-image (P2I)} branch and a \textit{point-to-point (P2P)} branch. See Fig.~\ref{fig:p2p}.

In the P2I branch, we train the point cloud encoder to predict the frozen image representation of a specific view using the 3D point cloud and the corresponding view index. Formally, given a point cloud \( p_i \) and a view index \( \gamma^t \), we compute its latent embedding using the point encoder \( \mathbf{P} \) and use the point-to-image predictor \( \mathbf{H}_{p2i} \) to generate the predicted image representation \( \hat{I}_i^{\gamma^t} \). We supervise it using the frozen image encoder \( \mathbf{I}_{\alpha} \) as:

\begin{equation}
\begin{split}
    \hat{I}_i^{\gamma^t} &= \mathbf{H}_{p_{2}i}(\mathbf{P_{\beta}}(p_i), \gamma^t)\\
    \mathcal{L}_{p_{2}i} &= \sum_i \sum_{t} \mathcal{L}(\mathbf{I}_{\alpha}(i^{\gamma^t}_i), \hat{I}_i^{\gamma^t})
\end{split}
\end{equation}

Here, \( i^{\gamma^t}_i \) denotes the image rendered from the point cloud \( p_i \) under view index \( \gamma^t \), and \( \mathcal{L} \) is the smooth L1 loss.

In the P2P branch, the goal is to predict the representation of a target point cloud view \( p_k \) from a source point cloud \( p_i \) using their relative view transformation \( \gamma^s \rightarrow \gamma^t \). The predicted latent \( \hat{P}_k \) is obtained by:

\begin{equation}
\begin{split}
    \hat{P}_k &= \mathbf{H}_{p2p}(\mathbf{P_{{\beta}_{ema}}}(p_i), \gamma^s \rightarrow \gamma^t)\\
    \mathcal{L}_{p_{2}p} &= \sum_k \sum_{s,t} \mathcal{L}(\mathbf{P_{\beta}}(p_k), \hat{P}_k)
\end{split}
\end{equation}

Here we have the learnable point encoder $\mathbf{P_{\beta}}$ and the weights of the model $\mathbf{P_{{\beta}_{ema}}}$ as the exponential moving average of $\mathbf{P_{\beta}}$ with a momentum of $\lambda$

\begin{equation} \label{eq:ema}
    \begin{split}
        \mathbf{P}_{\beta_{ema}} = \mathbf{P}_{\beta_{ema}} \times\lambda+\mathbf{P}_{\beta}\times(1-\lambda)
    \end{split}
\end{equation}

The final loss function combines both objectives:

\begin{equation}
    \mathcal{L} = \mathcal{L}_{p_{2}i} +\mathcal{L} _{p_{2}p}
\end{equation}

However, this approach results in an imbalance in the scales of the two loss terms $\mathcal{L}_{p_{2}i}$ and $\mathcal{L} _{p_{2}p}$; thus affects performance.

\subsection{Comparative Architectural Results}

\begin{table}[h]
    \centering
    \small
    \setlength{\tabcolsep}{2pt} 
    \renewcommand{\arraystretch}{1.3} 
    \resizebox{0.75\columnwidth}{!}{ 
    \begin{tabular}{l c c c} 
        \toprule
        & \textbf{$P_2I + I_2I$}& \textbf{$P_2I + P_2P$}& \textbf{$P_2I$}\\
        \midrule
        Image Learning& \checkmark & $\times$ & $\times$\\
        Point Teacher& $\times$ & \checkmark & $\times$\\
        \bottomrule
 \textbf{  Accuracy (\%)  }& 91.73 & 92.02 & \textbf{94.2}\\
        \bottomrule
    \end{tabular}
    }
    \caption{\textbf{Comparison of alternative CrossJEPA architectural variants.}
    We evaluate three candidate designs: (1) \textbf{$P_{2}I + I_{2}I$} which adds an auxiliary image-learning branch, (2) \textbf{$P_{2}I + P_{2}P$} which introduces a point-to-point predictive loss, and (3) the final \textbf{CrossJEPA ($P_{2}I$)} design.
    Both multi-branch variants suffer from instability, gradient imbalance, and higher memory and compute overhead, whereas the pure $P_{2}I$ configuration yields the best accuracy and the simplest, most efficient architecture.}
    \label{table:xjepa-architectures}
\end{table}

Table~\ref{table:xjepa-architectures} evaluates the aforementioned CrossJEPA-inspired architectural variants and motivates our final design choice.
The configuration \textbf{$P_{2}I + I_{2}I$}, which additionally learns in the image domain, frequently suffered from model collapse due to unstable gradients from the image-learning branch.
The \textbf{$P_{2}I + P_{2}P$} variant, which incorporates a point-to-point predictive loss, showed imbalanced optimization dynamics: the two loss terms $\mathcal{L}_{p{2}i}$ and $\mathcal{L}_{p{2}p}$ differ significantly in scale and produced inconsistent gradients, resulting in poor convergence.
Both of these multi-branch designs were also more memory intensive and substantially slower to train.

In contrast, the pure CrossJEPA configuration ($P_{2}I$), which predicts only from the point encoder to the frozen image teacher, achieves the highest performance (94.2\%), while being the most memory-efficient and computationally lightweight.
This streamlined design avoids unnecessary competing objectives, provides clean supervision from a strong 2D teacher, and leads to faster and more stable pretraining.
For these reasons, we adopt the $P_{2}I$ architecture as the core CrossJEPA framework and evaluated it in the main paper. 
Additional ablations of this CrossJEPA framework are detailed below.


\section{Further Ablations on Selected CrossJEPA ($P_2I$-only Setup)}

In each table, \textbf{Acc.} represents the ModelNet40 \cite{modelnet40} linear SVM accuracy in percentage, \textbf{Sat. Epoch} denotes the epoch at where the maximum accuracy was obtained, \textbf{Epoch T.} denotes the average time per epoch, \textbf{\#Params} denotes the total number of learnable parameters during pretraining, and \textbf{FLOPs} denotes the approximate floating point operation count per training sample \ie, for a single forward pass on a point cloud sample (if view count is 6 then it accounts for 1 point cloud forward pass and 6 predictions). 
Do note that all ablations are w.r.t. the baseline configuration described in Sec.~\ref{sec:methodology}.
\\

\begin{table}[h]
\centering
\caption{\textbf{Effect of point encoder and predictor size on performance and efficiency.} 
The large point encoder with a base predictor and the small point encoder with a large predictor achieve the highest accuracies, with the latter being more computationally efficient. Here, the FLOPs reflect a single forward pass for a single point cloud sample.}
\setlength{\tabcolsep}{2pt} 
\small  
\begin{tabular}{@{}lcccccc@{}}
\toprule
Point enc. & Predictor & Acc. (\%) & Sat. Epoch & \#Params & FLOPs \\
\midrule
small & small & 93.60 & 70 & 11.4M & 1.4B \\
small & base  & 93.68 & 59 & 14.1M & 1.5B \\
small & large & 93.96 & 102 & 16.7M & 1.7B \\
base  & small & 93.68 & 56 & 14.1M & 1.5B \\
base  & base  & 93.80 & 84 & 16.7M & 1.7B \\
base  & large & 93.48 & 58 & 19.4M & 1.8B \\
large & small & 93.40 & 44 & 16.7M & 1.7B \\
large & base  & 93.96 & 115 & 19.4M & 1.8B \\
large & large & 93.40 & 39 & 22.1M & 1.9B \\
\bottomrule
\end{tabular}
\label{tab:predictor_point_size}
\end{table}

\noindent\textbf{Predictor and Point Encoder Scaling.}
In the main experiments (Sec.~\ref{sec:ablations}), we scaled the predictor and point encoder sizes jointly (Table.~\ref{tab:model-scaling}) and demonstrated the scalability of CrossJEPA. 
Here, we decouple them and tune each component to study its individual effect on performance and efficiency, see Table.~\ref{tab:predictor_point_size}.
We interpret scaling each component as increasing its information capacity, where an appropriate balance between this capacity yields better performance and efficiency.
\\


\noindent\textbf{Pre-training Dataset.}
We pretrain CrossJEPA using ShapeNet \cite{shapenet} and a subset of Objaverse-XL \cite{objaverseXL}.
To assess data efficiency, we compare pretraining on ShapeNet alone versus the combined Objaverse + ShapeNet corpus (Table~\ref{tab:pretrain_dataset}).  
Training only on ShapeNet already yields a strong 93.27\% linear accuracy on ModelNet40, which is competitive with or better than many recent baselines \cite{act, crosspoint, i2p-mae}.  
Adding Objaverse modestly improves accuracy to 93.84\% while also reducing the number of required epochs from 161 to 71. 
This demonstrates that CrossJEPA benefits in performance and efficiency from additional data but does not rely on large pretraining sets to achieve competitive performance.
\\

\begin{table}[h]
\centering
\small
\setlength{\tabcolsep}{2pt}
\caption{\textbf{Effect of pretraining dataset.} 
CrossJEPA trained only on ShapeNet already achieves competitive performance compared with recent baselines, and adding a subset of Objaverse further improves accuracy while requiring fewer epochs. Here, average epoch duration (Epoch T.) is reported in mm:ss format.}
\label{tab:pretrain_dataset}
\begin{tabular}{lccc}
\toprule
Pretraining dataset & Acc. (\%) & Sat. Epoch & Epoch T. \\
\midrule
ShapeNet & 93.27     & 161    & 0:35          \\
Objaverse + subset of ShapeNet & 93.84  & 71 & 2:20          \\
\bottomrule
\end{tabular}
\end{table}


\noindent\textbf{Data Efficiency.}
CrossJEPA exhibits strong data efficiency when pretrained with only a fraction of the available Objaverse + subset of ShapeNet corpus (Table~\ref{tab:data_efficiency}). 
Even with just 20\% of the data, the model reaches a competitive 93.15\% linear accuracy on ModelNet40, only slightly below the 93.84\% obtained with the full dataset under the baseline configuration. 
This suggests that the frozen 2D teacher already possesses rich semantic mutual knowledge shared between images and point clouds, allowing the point encoder to learn effectively even when 3D data is scarce.
When the data fraction is small, convergence is slower, whereas with 100\% data convergence is faster, though the total pretraining time increases.
Overall, this behavior highlights the strength of cross-modal image-to-point knowledge transfer and indicates that CrossJEPA can mitigate the limitations of small 3D datasets, strengthening SSL in settings where large scale 3D collections are difficult to obtain.
\\

\begin{table}[t]
\centering
\setlength{\tabcolsep}{2pt}
\caption{\textbf{Data efficiency of CrossJEPA.} 
We evaluate CrossJEPA using different fractions of the pretraining set (Objaverse + subset of ShapeNet).
CrossJEPA maintains strong accuracy even with only 20\% of the data, demonstrating the effectiveness of image-to-point knowledge transfer in low data regimes.
This highlights the strong mutual information between 2D and 3D modalities and shows that CrossJEPA leverages the 2D teacher to compensate for limited 3D data. Here, average epoch duration (Epoch T.) is reported in mm:ss format.}
\begin{tabular}{cccc}
\toprule
Data fraction & Acc. (\%) & Sat. Epoch & Epoch T. \\
\midrule
20\%  & 93.15 & 132 & 00:39 \\
40\%  & 93.64 & 83  & 01:09 \\
60\%  & 93.31 & 33  & 01:40 \\
80\%  & 93.80 & 63  & 02:10 \\
100\% & 93.84 & 71  & 02:20 \\
\bottomrule
\end{tabular}
\label{tab:data_efficiency}
\end{table}


\noindent\textbf{Caching Mechanism.}
To quantify the benefit of our caching mechanism, we compare pretraining with and without cached image embeddings (Table~\ref{tab:caching}).
Without caching, the frozen ViT-B/14 teacher computes embeddings for all 24 rendered image views of size 518 of each point cloud object at every iteration, adding about 2.8T FLOPs per sample on top of the 1.5B FLOPs of the point encoder and predictor. This heavy overhead limits the batch size to 8 and results in roughly \emph{16 hours per epoch} on a single RTX 4090. 
With caching, we precompute these 24 view embeddings once and reuse them throughout training, so only the 1.5B trainable FLOPs remain. This allows us to increase the batch size to 256 and reduce epoch time to about \emph{2 minutes and 20 seconds}, while maintaining the same accuracy. 
Thus, caching is a byproduct of our frozen image teacher design choice that makes CrossJEPA practically efficient, potentially enabling large scale pretraining on standard hardware.
Our caching mechanism is extensively described in Sec.~\ref{sec:caching}.
\\

\begin{table}[t]
\centering
\caption{
\textbf{Effect of caching view embeddings on pretraining efficiency.}
We ablate this by comparing the performance when computing 24 image view embeddings for each point cloud sample using a frozen ViT-B/14 (518 image size) for every epoch versus using our caching mechanism, where all 24 view embeddings are precomputed once and reused across every epoch.
Our approach avoids 24 additional forward passes per sample (\(\approx 2.8\text{T}\) extra FLOPs without gradient tracking), so the trainable cost stays at \(\approx 1.5\text{B}\) FLOPs per sample for the
point encoder and predictor.
As a result, we can increase the batch size from 8 to 256 and reduce epoch time from 16 hours to about 2 minutes and 20 seconds on a single NVIDIA RTX 4090 GPU.
Here, FLOPs reflect a single forward pass for a single point cloud sample.
}
\label{tab:caching}
\setlength{\tabcolsep}{2pt}
\small  
\begin{tabular}{@{}lcccc@{}}
\toprule
Setting  & Epoch time & Batch size & FLOPs & Views \\
\midrule
With cache & 2m 20s & 256 & 1.5B (trainable) & 24 \\
No cache   & 16 h & 8 & 1.5B (trainable) + 2.8T & 24 \\
\bottomrule
\end{tabular}
\end{table}


\noindent\textbf{View Sampling Strategies.}
We next study how different view sampling strategies affect CrossJEPA.
We render 2D views of a 3D point cloud by moving and changing the orientation of the camera in a chosen coordinate system while keeping the point cloud fixed at the origin.
Table~\ref{tab:view_sampling} compares the sampling methods we explored.
We denote spherical sampling as changing the camera yaw and pitch, together with its radius from the origin; cylindrical sampling as changing the camera yaw, height, and radial distance; and Cartesian sampling as changing the camera position along the $x$, $y$, and $z$ axes.
Our default strategy follows the spherical parameterization.
All three parameterizations achieve very similar ModelNet40 linear accuracies, around 93.8\%, with only minor differences in the number of epochs required for convergence (cylindrical being slightly faster).
This indicates that CrossJEPA is largely insensitive to the choice of coordinate system for view sampling, allowing practitioners to select whichever parameterization best fits their rendering pipeline.

\begin{table}[t]
\centering
\small
\setlength{\tabcolsep}{2pt}
\caption{\textbf{Effect of different view sampling parametrization on CrossJEPA.} 
Comparison of spherical, cylindrical, and Cartesian camera parameterizations for view sampling. 
All three strategies yield similar ModelNet40 linear accuracies, indicating that CrossJEPA is robust to the choice of parametrization for generating rendered views.
Here, $r$ denotes the radial distance from the origin, $h$ denotes the axial distance (height) from the $x$–$y$ plane, and $x$, $y$, and $z$ denote the camera position along the respective axes.}
\label{tab:view_sampling}
\begin{tabular}{l l c c c}
\toprule
Sampling & Parameterization      & Acc. (\%) & Sat. Epoch & Epoch T. \\
\midrule
Spherical  & $(\text{yaw}, \text{pitch}, r)$ & 93.84 & 71 & 02:20\\
Cylindrical & $(\text{yaw}, h, r)$ & 93.88 & 53 & 02:20 \\
Cartesian  & $(x, y, z)$ & 93.88 & 83 & 02:20\\
\bottomrule
\end{tabular}
\end{table}






\begin{figure*}[h]
    \centering
    \includegraphics[width=\textwidth]{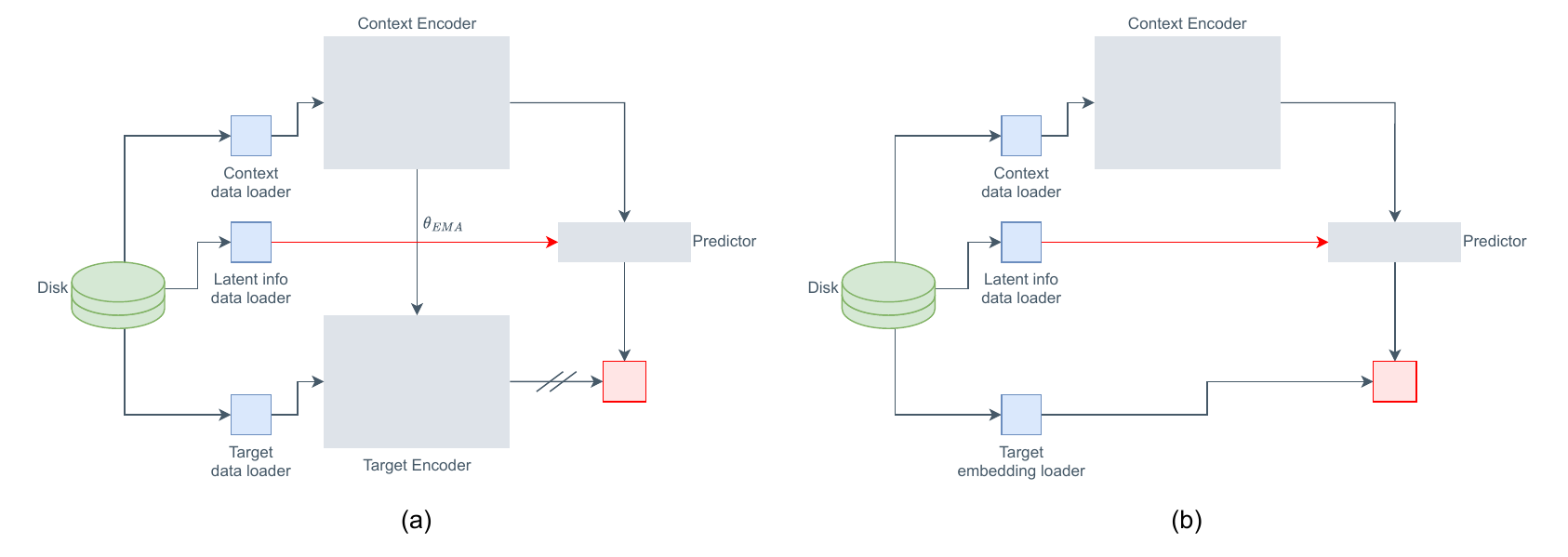}
    \caption{
    \textbf{Standard vs. Cached JEPA Pretraining Architecture Pipelines.}
    \textbf{(a)} Standard JEPA-style training with an online/EMA target encoder: every step computes target features, incurring continual target-branch forward passes. \textbf{(b)} CrossJEPA with a frozen image encoder and \emph{cached} target embeddings: targets are precomputed and streamed from disk, removing the target-branch compute during training and yielding \emph{amortized} speedups across epochs.}
    \label{fig:xjepa-caching}
\end{figure*}

\section{Caching Mechanism}
\label{sec:caching}

\textcolor{black}{To optimize runtime performance and reduce redundant computation, the \texttt{cache\_embeddings} method precomputes and stores image embeddings for each view in the dataset using the selected frozen image encoder. By doing so, the embeddings—once calculated—are saved in an HDF5 cache file (.h5), indexed by their directory names. This avoids recomputing embeddings for the same input images during each run, significantly reducing GPU memory usage and inference latency. Since the encoder remains frozen during training and inference, the precomputed embeddings remain valid throughout, making this an effective one-time preprocessing step. This design choice ensures faster training and evaluation cycles, especially in large-scale multimodal learning tasks where image encoders are typically compute-intensive.}

\textcolor{black}{The caching mechanism is seamlessly integrated into the dataset initialization phase, allowing the model to load only the required embeddings into memory without needing to forward entire image batches through the encoder. The use of HDF5 with dynamic dataset resizing and chunking allows for scalable and efficient I/O operations, even for thousands of view-specific image embeddings. The \texttt{flush()} mechanism ensures that memory is not overloaded during batch processing and that incomplete embedding groups are skipped, preserving dataset integrity.}

\begin{lstlisting}[
    style=mystyle,
    caption={\textcolor{black}{The \texttt{cache\_embeddings} method of our \texttt{Dataset} class, which is called during initialization to compute and cache image embeddings for all views. This one-time image embedding caching mechanism significantly accelerates pre-training and reduces redundant GPU memory load.}},
    label={lst:caching_code},
    basicstyle=\ttfamily\scriptsize
]
from typing import Callable, Dict
from os.path import dirname, exists as ospe
import os

import h5py
import numpy as np
from tqdm import tqdm
import torch
from torch.utils.data import DataLoader

from src.datasets.shapenet_image import ShapeNetImage
from src.utils.config import make_obj_from_conf


@torch.no_grad()
def cache_embeddings(
    self, img_model_conf: Dict, image_trans: Callable[[torch.Tensor], torch.Tensor]
) -> Dict[str, int]:
    """
    Get the image embeddings for all the images in the dataset using the specified image model.
    The embeddings are cached. If the embeddings are already cached, they will be skipped.
    """

    if ospe(self.em_cache_path):
        with h5py.File(self.em_cache_path, "r") as f:
            available_emb_img_dirs = {s.decode("utf-8") for s in f["img_dirs"][()]}
            missing_emb_img_dirs = [
                d for d in self.img_dirs if d not in available_emb_img_dirs
            ]
    else:
        missing_emb_img_dirs = self.img_dirs

    if len(missing_emb_img_dirs) > 0:
        os.makedirs(dirname(self.em_cache_path), exist_ok=True)
        missing_emb_img_paths = [
            f"{dir_}/{p}" for dir_ in missing_emb_img_dirs for p in os.listdir(dir_)
        ]
        ds = ShapeNetImage(
            image_trans, "full", return_val="path", img_paths=missing_emb_img_paths
        )
        model = make_obj_from_conf(img_model_conf).cuda().eval()
        dl = DataLoader(ds, batch_size=64, shuffle=False, num_workers=8)
        mode = "a" if ospe(self.em_cache_path) else "w"
        flush_freq = 100
        with h5py.File(self.em_cache_path, mode) as f:
            calc_embs = {}
            if "img_dirs" not in f:
                f.create_dataset(
                    "img_dirs", shape=(0,), maxshape=(None,), dtype="S", chunks=True
                )
            if "embeddings" not in f:
                f.create_dataset(
                    "embeddings",
                    shape=(0, 36, 768),
                    maxshape=(None, 36, 768),
                    dtype="f4",
                    chunks=True,
                )

            def flush():
                in_memory_dirs = list(calc_embs.keys())
                for p in in_memory_dirs:
                    emb, mask = calc_embs[p]
                    if mask.sum() == 36:
                        f["embeddings"].resize(
                            (f["embeddings"].shape[0] + 1, 36, emb.shape[1])
                        )
                        f["embeddings"][-1] = emb
                        f["img_dirs"].resize((f["img_dirs"].shape[0] + 1,))
                        f["img_dirs"][-1] = p.encode("utf-8")
                        calc_embs.pop(p)

            for i, (imgs, paths) in enumerate(
                tqdm(
                    dl,
                    desc=f"Computing image embeddings for model {img_model_conf.params.backbone}",
                )
            ):
                embs_ = model(imgs.cuda()).cpu()
                for emb, path in zip(embs_, paths):
                    dirnm, basenm = os.path.split(path)
                    yaw, pitch = os.path.splitext(basenm)[0].split("_")
                    idx = int(int(yaw) * 6 + int(pitch))
                    if dirnm not in calc_embs:
                        calc_embs[dirnm] = (
                            np.empty((36, embs_.shape[1]), dtype="f4"),
                            np.zeros(36, dtype=int),
                        )
                    calc_embs[dirnm][0][idx] = emb.numpy()
                    calc_embs[dirnm][1][idx] = 1
                if i % flush_freq == 0:
                    flush()
            flush()
            assert (
                len(calc_embs) == 0
            ), f"Some embeddings could not be computed fully: {list(calc_embs.keys())}"
    with h5py.File(self.em_cache_path, "r") as f:
        cached_img_dirs = [s.decode("utf-8") for s in f["img_dirs"][()]]
    return dict(zip(cached_img_dirs, range(len(cached_img_dirs))))
\end{lstlisting}

\section{Analysis of CrossJEPA from an Information Theoretic Perspective} 
\label{sec:theoretical_justification}

There have been recent attempts to explain the superior performance of self supervised methods using information theory \cite{lecunn_information_theory}. 
However, to the best of our knowledge, no work has analyzed the performance of JEPA architectures from this perspective. 
We present a basic framework to explain multimodal JEPA architectures using information theory \cite{information_theory_textbook}.

\subsection{Pose Conditioning as Nuisance Suppression}
\paragraph{Setup.}
Let $S$ denote the pose-invariant content (represented by a point cloud), $P$ the pose (a nuisance for point-cloud tasks), and $Y$ the supervision target. For CrossJEPA, $Y = I_\alpha(\gamma)$, the image embedding at pose $\gamma$. Ideally, the supervision signal should depend only on the object's intrinsic structure $S$, not on pose $P$. While frozen image embeddings $Y$ are not strictly pose-invariant, conditioning the predictor on $P$ effectively enforces the independence $Y \!\perp\! P \mid S$ from the point encoder's perspective \cite{achille2018emergence,achille2018information,alain2016understanding}. The point encoder is $Z_\theta = f_\theta(\text{points}(S))$, the predictor is $h_\phi(Z_\theta,P)$, and training uses per-sample loss $\ell(h_\phi(Z_\theta,P),Y)$ with stochastic gradient $g=\nabla_\theta \ell$.

\paragraph{Assumptions.}
(A1) $E[Y\mid S,P]=m(S)$, \ie the conditional mean depends only on $S$ \cite{achille2018emergence}.  
(A2) $\ell$ is differentiable and expectations commute with $\nabla_\theta$ \cite{bottou2018optimization}.

\paragraph{Proposition 1 (Pose decorrelation).}
If the predictor is additive $h_\phi(z,p)=a_\phi(z)+b_\phi(p)$ and the loss is Smooth L1,
\[
E[\nabla_z \ell(h_\phi(Z_\theta,P),Y)\mid S,P]
=E[\nabla_z \ell(h_\phi(Z_\theta,P),Y)\mid S].
\]
For clarity, we illustrate the gradient-decorrelation effect under an additive predictor $h_\phi(z,p)=a_\phi(z)+b_\phi(p)$. While our actual predictor is more general (MLP over concatenated inputs), the same intuition applies: conditioning on $p$ ensures that pose variability is absorbed by the predictor, reducing its influence on encoder gradients \cite{achille2018information,tschannen2020self}.\\
\textit{Proof.}
For Smooth L1 with transition $\beta>0$,
\[
\frac{\partial \ell}{\partial u} =
\begin{cases}
(u-y)/\beta, & |u-y|<\beta,\\
\text{sign}(u-y), & \text{otherwise},
\end{cases}
\]
with $u=h_\phi(z,p)$. Hence $\nabla_z \ell=\nabla_z a_\phi(z)\,\frac{\partial \ell}{\partial u}$.  
Taking $E[\cdot \mid S,P]$ and using $E[Y\mid S,P]=m(S)$, the conditional expectation depends only on $S$. The same holds for $E[\cdot \mid S]$, proving equality. \hfill $\square$
\begin{figure*}[] 
    \centering
    \includegraphics[width=0.9\textwidth]{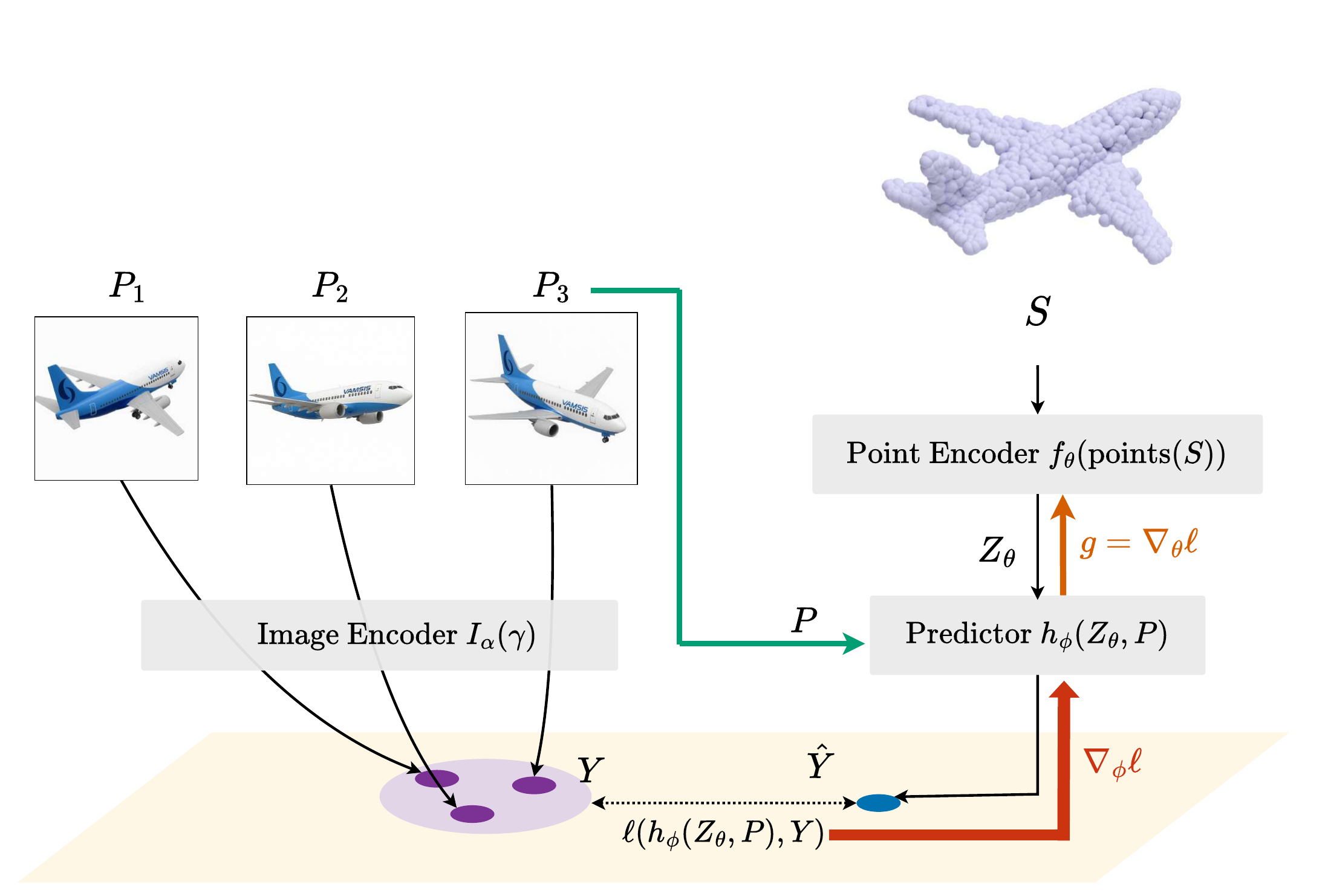} 
\caption{\textbf{Pose conditioning as nuisance suppression.} 
  The point cloud $S$ is encoded into a latent $Z_\theta$, which together with pose $P$ is fed into the predictor $h_\phi$ to produce $\hat Y$ in the image-embedding space. 
  The frozen image embeddings $Y=I_\alpha(\gamma)$ provide supervision via a Smooth L1 loss. 
  Gradients flow to both the predictor $(\nabla_\phi \ell)$ and the encoder $(g=\nabla_\theta \ell)$. 
  Conditioning on pose ensures that nuisance variation is absorbed at the predictor, preventing pose-dependent gradient leakage into the encoder and yielding lower-variance, pose-invariant features (Proposition~1--Theorem~2).}
    \label{fig:info_theory_nuisance_suppression}
\end{figure*}
\paragraph{Theorem 1 (Variance reduction).}
By the law of total variance,
\[
\mathrm{Var}(g) = E[\mathrm{Var}(g\mid S)] + \mathrm{Var}(E[g\mid S]) 
\;\ge\; \mathrm{Var}(E[g\mid S]).
\]
\textit{Proof.}  
This is the standard variance decomposition \cite{bottou2018optimization}. By Proposition~1, $E[g\mid S,P]=E[g\mid S]$, so pose-induced fluctuations only contribute to the inner variance term, which vanishes when $P$ is conditioned. This mirrors findings in variance reduction for SGD \cite{zhang2020variance,robbins1951stochastic}. \hfill $\square$

\paragraph{Proposition 2 (Risk preservation).}
Let $R(\theta,\phi)=E[\ell(h_\phi(Z_\theta,P),Y)]$ and define the marginalized loss
\[
\bar\ell(z)=E[\ell(h_\phi(z,P),Y)\mid S].
\]
Then
\[
\nabla_\theta R(\theta,\phi)=E[\nabla_\theta \bar\ell(Z_\theta)].
\]
\textit{Proof.}  
By definition,
\[
R(\theta,\phi) = E[\ell(h_\phi(Z_\theta,P),Y)].
\]
Differentiating under the expectation (Assumption A2),
\[
\nabla_\theta R(\theta,\phi) = E[\nabla_\theta \ell(h_\phi(Z_\theta,P),Y)].
\]
Now expand the outer expectation via the law of iterated expectations:
\[
E[\nabla_\theta \ell(h_\phi(Z_\theta,P),Y)] 
= E\!\big[\, E[\nabla_\theta \ell(h_\phi(Z_\theta,P),Y)\mid S,P] \,\big].
\]
By Proposition~1, the inner conditional expectation is invariant to $P$, i.e.
\[
E[\nabla_\theta \ell(h_\phi(Z_\theta,P),Y)\mid S,P] 
= E[\nabla_\theta \ell(h_\phi(Z_\theta,P),Y)\mid S].
\]
But the right-hand side is precisely the gradient of the marginalized loss,
\[
E[\nabla_\theta \ell(h_\phi(Z_\theta,P),Y)\mid S] = \nabla_\theta \bar\ell(Z_\theta).
\]
Therefore,
\[
\nabla_\theta R(\theta,\phi) = E[\nabla_\theta \bar\ell(Z_\theta)],
\]
which shows that the population risk gradient coincides with the gradient of the pose-marginalized objective \cite{alain2016understanding}. \hfill $\square$

\paragraph{Theorem 2 (Invariance pressure).}
If $h_\phi$ is expressive in $P$ and the encoder is regularized (e.g.\ weight decay), then any Bayes-optimal solution can be chosen such that
\[
I(Z_\theta;P) \;\text{is minimal among risk-equivalent encoders.}
\]
\textit{Proof.}  
Let $\mathcal{F}$ denote the set of encoder--predictor pairs $(f_\theta,h_\phi)$ that achieve the Bayes risk for the task. By Assumption~A1, the supervision target satisfies $Y \perp P \mid S$ \cite{achille2018emergence,tishby1999information}, so once $S$ is known, $P$ carries no additional information about $Y$. Because the predictor $h_\phi$ has explicit access to $P$, any dependence of the encoder representation $Z_\theta$ on $P$ is redundant for achieving the optimal risk \cite{shwartz2017opening}.  

Formally, for any $(f_\theta,h_\phi) \in \mathcal{F}$, construct a new encoder $\tilde f$ such that $\tilde f(S)$ is the conditional expectation of $Z_\theta$ given $S$. This $\tilde f$ yields a representation $\tilde Z$ satisfying $I(\tilde Z;P)=0$, while preserving the conditional distribution needed to achieve the same Bayes risk.  

Among all risk-equivalent encoders, regularization (e.g.\ weight decay or information bottleneck penalties) favors solutions of minimal complexity \cite{tishby1999information,alemi2017deep}. This implies selecting an encoder that discards superfluous dependence on $P$, i.e.\ one that minimizes $I(Z_\theta;P)$. Hence, any Bayes-optimal solution can be chosen such that
\[
I(Z_\theta;P) \;\text{is minimal among risk-equivalent encoders.}
\]\hfill $\square$

\paragraph{Consequences.}
Proposition~1 shows pose-conditioning removes nuisance-induced correlation from encoder gradients. Theorem~1 guarantees reduced gradient variance, improving sample efficiency. Proposition~2 confirms that the true population risk gradient is preserved, while Theorem~2 formalizes the invariance pressure: the encoder naturally drops pose information once the predictor accounts for it.

 Conditioning the predictor on additional image-specific but point-irrelevant features further improves predictive accuracy. This follows directly from the same line of argument: incorporating such side information reduces uncertainty in the target space without altering the intrinsic structure captured by the point encoder. In our experiments, we observe consistent gains in accuracy when providing pose information and image color cues. Quantitative results are summarized in $\ref{table:supp_color+pose}$.

\begin{table}[h]
    \centering
        \caption{\textbf{Effect of latent information for predictor in JEPA:} Adding camera pose and color histogram information consistently improves both accuracy and convergence speed. The best configuration yields 94.0\% accuracy with fewer epochs, confirming that CrossJEPA leverages latent cues to accelerate learning.}
    \renewcommand{\arraystretch}{1.3}
    \resizebox{0.9\columnwidth}{!}{
    \begin{tabular}{lcc} 
        \toprule
        \textbf{Experiment} & \textbf{Acc.} & \textbf{Sat. Epoch} \\ 
        \midrule
        No Latent Info & 93.68\% & 93 \\
        Cam Pose & 93.84\% & 71 \\
        Cam Pose + Color Hist.  & 94.04\% & 67 \\
        \bottomrule
    \end{tabular}
    }
    \label{table:supp_color+pose}
\end{table}

\subsection{Inefficiencies in Reconstruction for Representation Learning}

\paragraph{Setup.}
We denote by $r(S,P)$ the deterministic rendering function that maps the structure $S$ under pose $P$ to a coarse observation (geometry + viewpoint without fine detail). The full observation is
\[
X \;=\; r(S,P) + H,
\]
where $H$ captures \emph{fine-grained, task-irrelevant details} (e.g., textures, micro-geometry, small appearance variations). 
We model $H$ as a residual detail term with $E[H\mid S,P]=0$; when stated explicitly, we also assume $H \perp (S,P)$ for variance decompositions \cite{vincent2010stacked}. 
A reconstruction objective trains a decoder $d_\psi$ via
\[
\mathcal L_{\mathrm{rec}}(\theta,\psi) \;=\; E\big[\|X - d_\psi(Z_\theta)\|^2\big].
\]

\paragraph{Lemma 1 (MMSE decoder).}
For any fixed encoder $Z_\theta$, the MSE–optimal decoder is
\[
d_\psi^\star(z) \;=\; E[X \mid Z_\theta = z],
\]
and the optimal reconstruction risk equals
\[
\begin{aligned}
    \mathcal L_{\mathrm{rec}}(\theta,\psi^\star) \;=\; E\!\left[\mathrm{Var}(X \mid Z_\theta)\right].
\end{aligned}
\]
\textit{Proof.} Standard regression result: the minimizer of $E[\|X-\hat X(Z_\theta)\|^2]$ is $\hat X=E[X\mid Z_\theta]$, with minimum $E[\mathrm{Var}(X\mid Z_\theta)]$ \cite{bishop2006pattern}. \hfill $\square$

\paragraph{Proposition 3 (Irreducible reconstruction term from fine detail).}
In general,
\begin{align*}
\mathcal L_{\mathrm{rec}}(\theta,\psi^\star)
&= E\!\left[\mathrm{Var}\!\big(r(S,P)\mid Z_\theta\big)\right] 
 + E\!\left[\mathrm{Var}\!\big(H\mid Z_\theta\big)\right] \\
&\quad + 2\,E\!\left[\mathrm{Cov}\!\big(r(S,P),H \mid Z_\theta\big)\right].
\end{align*}
If, in addition, $H \perp (S,P)$ (hence $H \perp Z_\theta$), this simplifies to
\begin{align*}
\mathcal L_{\mathrm{rec}}(\theta,\psi^\star)
= E\!\left[\mathrm{Var}\!\big(r(S,P)\mid Z_\theta\big)\right] \;+\; \mathrm{Var}(H).
\end{align*}
Since $Z_\theta=f_\theta(S)$ is a (possibly lossy) function of $S$,
\begin{align*}
E\!\left[\mathrm{Var}\!\big(r(S,P)\mid Z_\theta\big)\right]
\;\ge\; E\!\left[\mathrm{Var}\!\big(r(S,P)\mid S\big)\right],
\end{align*}
with equality iff $Z_\theta$ is sufficient for $S$.

\noindent\textit{Proof.} 
By Lemma~1,
\begin{align*}
\mathcal L_{\mathrm{rec}}(\theta,\psi^\star)
= E\!\left[\mathrm{Var}(X \mid Z_\theta)\right],
\qquad X = r(S,P)+H.
\end{align*}
For any square–integrable $U,V$ and $\sigma$–algebra $\mathcal G$,
\begin{align*}
\mathrm{Var}(U+V \mid \mathcal G)
= &\mathrm{Var}(U \mid \mathcal G)
 + \mathrm{Var}(V \mid \mathcal G)\\
 +& 2\,\mathrm{Cov}(U,V \mid \mathcal G).
\end{align*}
Applying this with $U=r(S,P)$, $V=H$, and $\mathcal G=\sigma(Z_\theta)$ gives
\begin{align*}
\mathrm{Var}(X \mid Z_\theta)
&= \mathrm{Var}\!\big(r(S,P) \mid Z_\theta\big)
 + \mathrm{Var}\!\big(H \mid Z_\theta\big) \\
&\quad + 2\,\mathrm{Cov}\!\big(r(S,P), H \mid Z_\theta\big).
\end{align*}
Taking expectations yields the first displayed identity.

If $H \perp (S,P)$, then $H \perp Z_\theta$ (since $Z_\theta=f_\theta(S)$), so
\begin{align*}
\mathrm{Var}(H \mid Z_\theta)=\mathrm{Var}(H),
\qquad
\mathrm{Cov}\!\big(r(S,P),H \mid Z_\theta\big)=0,
\end{align*}
which gives the simplified form.

Finally, since $\sigma(Z_\theta)\subseteq\sigma(S)$, the $L^2$ projection property implies
\begin{align*}
E\!\left[\mathrm{Var}(W\mid S)\right]
\;\le\; E\!\left[\mathrm{Var}(W\mid Z_\theta)\right],
\end{align*}
for any square–integrable $W$. Equality holds iff $Z_\theta$ is sufficient for $S$ with respect to predicting $r(S,P)$ \cite{tsybakov2009nonparametric}. \hfill $\square$

\paragraph{Consequence.}
Driving $\mathcal L_{\mathrm{rec}}$ down forces $Z_\theta$ to encode \emph{predictable} components of $X$ arising from $P$ and from fine detail $H$, even when those are irrelevant to the supervision $Y$ \cite{bengio2013representation}.

\paragraph{Theorem 3 (Rate–distortion lower bound forces detail/nuisance retention).}
For any encoder $Z_\theta$ and decoder achieving $E[\|X-d_\psi(Z_\theta)\|^2]\le D$,
\[
I(Z_\theta;X) \;\ge\; R_X(D),
\]
where $R_X(D)$ is the rate–distortion function of $X$ under squared error \cite{cover2006elements}. 
Since
\[
I(Z_\theta;X)=I(Z_\theta;S)+I(Z_\theta;P,H\mid S),
\]
any sufficiently small distortion $D$ enforces
\[
I(Z_\theta;P,H\mid S) \;\ge\; R_X(D) - I(Z_\theta;S).
\]

\noindent\textit{Proof.}
Let $\widehat X := d_\psi(Z_\theta)$ be the reconstruction produced by the encoder--decoder pair.
Since $\widehat X$ is a deterministic function of $Z_\theta$, we have the Markov chain
$X \to Z_\theta \to \widehat X$.
By the data processing inequality,
\[
I(X;Z_\theta) \;\ge\; I(X;\widehat X).
\]
If $E[\|X-\widehat X\|^2]\le D$, then by the definition of the rate--distortion function under squared error,
\[
I(X;\widehat X) \;\ge\; R_X(D).
\]
Combining the two displays gives $I(X;Z_\theta)\ge R_X(D)$ \cite{cover2006elements}. \hfill $\square$

\paragraph{Theorem 4 (Gradient variance inflation under reconstruction).}
Let $\varepsilon := X - d_\psi(Z_\theta)$ be the residual and 
$J_\theta := \partial Z_\theta/\partial \theta$ the encoder sensitivity 
(absorbing decoder Jacobians if desired). For squared loss,
\[
\nabla_\theta \mathcal L_{\mathrm{rec}} = E[J_\theta^\top \varepsilon],
\quad
\mathrm{Cov}(\nabla_\theta \|\varepsilon\|^2 \mid S) 
= J_\theta^\top\, \mathrm{Cov}(\varepsilon\mid S)\, J_\theta .
\]
Decomposing the residual,
\[
\varepsilon 
= \big(r(S,P)-E[r(S,P)\mid Z_\theta]\big) 
+ \big(H - E[H\mid Z_\theta]\big),
\]
gives
\begin{align*}
&\mathrm{Cov}(\varepsilon\mid S)
= \mathrm{Cov}\!\big(r(S,P)-E[r(S,P)\mid Z_\theta] \mid S\big) \\
& + \mathrm{Cov}\!\big(H - E[H\mid Z_\theta]\big) \\
& + 2\,\mathrm{Cov}\!\big(r(S,P)-E[r(S,P)\mid Z_\theta],\, H - E[H\mid Z_\theta] \mid S\big).
\end{align*}
If $H \perp (S,P)$, the last two terms simplify to $\mathrm{Cov}(H)$ and $0$, 
so fine detail imposes an additive variance floor. Any residual dependence further increases this covariance. 
Hence both predictable pose variability and fine detail inflate gradient variance, slowing optimization \cite{bottou2018optimization,hardt2016train}. 
If prediction targets are instead defined after marginalizing over $P$ (as in our pose–conditioned predictor), 
the first covariance term strictly decreases whenever $r(S,P)$ varies with $P$.

\noindent\textit{Proof.}
Let $D_\psi(Z_\theta):=\partial d_\psi(Z_\theta)/\partial Z$ and 
$G_\theta:=\partial Z_\theta/\partial \theta$. 
For the per-sample squared loss $\|\varepsilon\|^2=\|X-d_\psi(Z_\theta)\|^2$,
\begin{align*}
\nabla_\theta \|\varepsilon\|^2
&= 2\,(\partial_\theta d_\psi(Z_\theta))^\top (d_\psi(Z_\theta)-X) \\
&= 2\,G_\theta^\top D_\psi(Z_\theta)^\top \varepsilon .
\end{align*}
Absorb constants and $D_\psi$ into 
$J_\theta^\top:=-\,2\,G_\theta^\top D_\psi(Z_\theta)^\top$, 
so that 
\[
\nabla_\theta \|\varepsilon\|^2 = J_\theta^\top \varepsilon,
\qquad
\nabla_\theta \mathcal L_{\mathrm{rec}} = E[J_\theta^\top \varepsilon].
\]

Since $J_\theta$ is measurable w.r.t.\ $S$,
\[
\mathrm{Cov}(J_\theta^\top U \mid S)
= J_\theta^\top\, \mathrm{Cov}(U\mid S)\, J_\theta,
\]
so with $U=\varepsilon$,
\[
\mathrm{Cov}(\nabla_\theta \|\varepsilon\|^2 \mid S) 
= J_\theta^\top\, \mathrm{Cov}(\varepsilon\mid S)\, J_\theta .
\]

Decompose
\[
\varepsilon
= \underbrace{r(S,P)-E[r(S,P)\mid Z_\theta]}_{\text{pose part}}
+ \underbrace{H-E[H\mid Z_\theta]}_{\text{detail part}} .
\]
Then
\begin{align*}
&\mathrm{Cov}(\varepsilon\mid S)
= \mathrm{Cov}\!\big(r(S,P)-E[r(S,P)\mid Z_\theta]\mid S\big) \\
& + \mathrm{Cov}\!\big(H - E[H\mid Z_\theta]\big) \\
& + 2\,\mathrm{Cov}\!\big(r(S,P)-E[r(S,P)\mid Z_\theta],\, H - E[H\mid Z_\theta]\mid S\big).
\end{align*}
If $H \perp (S,P)$ and $E[H]=0$, then $H \perp Z_\theta$, so 
\[
\mathrm{Cov}(H - E[H\mid Z_\theta])=\mathrm{Cov}(H),
\qquad
\mathrm{Cov}(\cdot,\cdot\mid S)=0.
\]
Thus fine detail contributes an additive covariance floor.

Finally, let $U=r(S,P)$ and $U_S=E[U\mid S]$. Then
\begin{align*}
&E[\|U-E[U\mid Z_\theta]\|^2 \mid S]
= E[\mathrm{Var}(U\mid Z_\theta,S)\mid S] \\
& + \mathrm{Var}(E[U\mid S]-E[U\mid Z_\theta]\mid S) \\
&\ge \mathrm{Var}(U_S - E[U_S\mid Z_\theta]\mid S).
\end{align*}
The inequality is strict whenever $U$ varies with $P$ given $S$. 
Hence replacing $U$ by $U_S$ (pose–marginal target) strictly reduces 
the leading covariance term, proving gradient variance reduction \cite{hardt2016train,zhang2020variance}. \hfill $\square$

\paragraph{Synthesis.}
Proposition~3 shows an \emph{irreducible} reconstruction term due to pose variability and fine detail; Theorem~3 formalizes that low distortion on $X$ demands high $I(Z_\theta;X)$, typically increasing $I(Z_\theta;P,H\mid S)$; Theorem~4 explains the optimization penalty via inflated gradient variance. In contrast, objectives that predict only task–relevant signals (our pose–conditioned predictor) avoid encoding fine-grained detail and yield lower-variance gradients.

\section{Analysis on Predictive Coding, Human Cognition, and CrossJEPA}

\label{sec:predictive_coding}

\paragraph{Motivation.}
Predictive coding (PC) proposes that cortex implements analysis-by-synthesis: higher areas generate top–down predictions of sensory input, lower areas compute \emph{prediction errors}, and these errors update internal representations and synapses \citep{RaoBallard1999,Friston2010,Clark2013,Bastos2012}. We view CrossJEPA as an engineered instance of this principle across modalities: a point-cloud representation (latent \emph{cause}) is used to \emph{predict} an image embedding for a queried 2D view (sensory \emph{consequence}); the discrepancy serves as the learning signal, in line with JEPA/I-JEPA’s \cite{lecun_a_path_towards, ijepa} error-minimization view of representation learning \citep{ijepa}.

\paragraph{Setup and notation.}
Let $S$ denote object identity/shape (intrinsic structure), $P$ a control variable (pose/viewpoint), and $Y$ the target in image-embedding space. The point encoder yields $Z_\theta=f_\theta(\text{points}(S))$. Conditioned on a view index $\gamma$ (instantiating $P$), the predictor outputs $\hat{Y}=h_\phi(Z_\theta,\gamma)$. A frozen image encoder $I_\alpha$ returns $Y=I_\alpha(i_\gamma)$ from the rendered/real view $i_\gamma$. Training minimizes $L(\hat{Y},Y)$ (e.g., $\ell_1$) and updates $(\theta,\phi)$ by the prediction error.

\begin{figure}[t]
    \centering
    \includegraphics[width=0.5\textwidth]{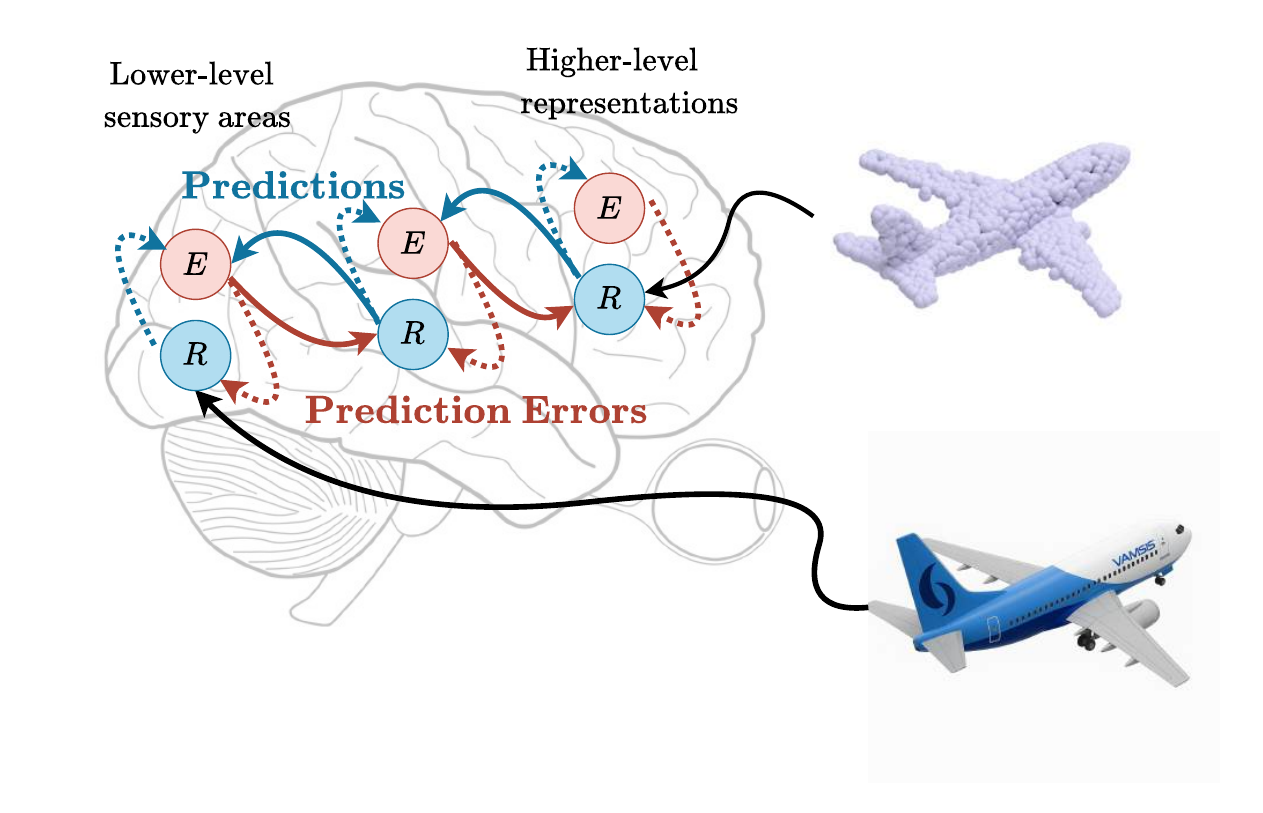}
    \caption{\textbf{Predictive coding in the brain and its analogy to CrossJEPA.} As described in the \emph{Supplementary} (Sec. \ref{sec:predictive_coding}), higher-level areas generate 
top--down predictions (blue arrows) of expected sensory input, while lower-level areas transmit 
bottom--up prediction errors (red arrows) when reality diverges from expectation. The observed 
plane (sensory input) is compared against the expected plane (internal prediction), and mismatches 
update higher-level representations. CrossJEPA follows the same principle: a 3D point cloud (latent 
cause) is used to predict 2D image embeddings (sensory consequences), with prediction errors 
driving the learning of object-centered representations.}
    \label{fig:predictive_coding}
\end{figure}

\paragraph{PC $\leftrightarrow$ CrossJEPA: alignment.}
Top–down prediction $\hat{Y}=h_\phi(Z_\theta,\gamma)$ corresponds to cortical predictions; bottom–up error $e=Y-\hat{Y}$ drives weight changes; conditioning by $\gamma$ plays the role of precision/attention by selecting which consequences to predict \citep{Bastos2012}. Stacking JEPA blocks yields hierarchical representations, mirroring PC’s multi-level organization \citep{Friston2010,Clark2013}.

\paragraph{Normative view.}
Under PC, the system maintains a predictive model from latent causes $(S,P)$ to observations. The optimal predictor approximates
\[
m(S,P)=E\!\left[ Y \mid S,P \right],
\]
so minimizing $E\|Y-h_\phi(Z_\theta,\gamma)\|$ implements error-driven learning \citep{Friston2010}. If $E[Y\mid S,P]=m(S)$ once $\gamma$ is provided—informally, $Y \perp P \mid S$ given the view—then $h_\phi$ learns an object-centered mapping. CrossJEPA formalizes this with targets in an embedding space, avoiding pixel-level generative burdens while preserving semantics \citep{ijepa}.

\paragraph{Why cross-modal prediction helps.}
PC emphasizes predicting lawful \emph{sensory consequences} of latent causes. In CrossJEPA, the larger context space (3D points) induces a family of lawful 2D consequences. Predicting into a frozen image-embedding manifold exploits a stable, semantically organized target and discourages shortcuts—consistent with analysis-by-synthesis accounts of perception \citep{Clark2013}.

\paragraph{Consequences and testable predictions.}
(1) \emph{View-consistency:} errors $e(\gamma)$ for a fixed $S$ should correlate across $\gamma$ more than across identities.  
(2) \emph{Precision effects:} per-dimension error reweighting (\eg Mahalanobis) should mimic PC precision, improving data efficiency \citep{Bastos2012}.  
(3) \emph{Analysis-by-synthesis ablation:} removing or corrupting $\gamma$ raises error; enriching controllable queries (lighting, focal length) tightens invariances in $Z_\theta$.  
(4) \emph{OOD poses:} predictors fall back to global shape regularities—lower error on coarse geometry, higher on fine texture—reflecting PC’s bias toward learned causes \citep{Clark2013}.

\paragraph{Relation to masking-based JEPA.}
Masking defines an information subspace within a modality; CrossJEPA uses a \emph{geometrically induced} subspace: 2D consequences of a 3D cause. Both fit PC (predict missing consequences), but cross-modal prediction avoids training-time mutilations and better matches real sensorimotor contingencies \citep{ijepa}.

\paragraph{Summary.}
Viewed through predictive coding, CrossJEPA operationalizes human-like analysis-by-synthesis: infer a cause-centric representation from points, predict lawful sensory consequences in image-embedding space, and learn solely from the mismatch \cite{RaoBallard1999,ijepa}.

\section{Learning from Point Clouds in the Wild}

Our proposed CrossJEPA method uses 3D point clouds along with their corresponding 2D visualizations with known camera projection parameters. The work demonstrates the proposed method for CAD models \cite{shapenet}, placing the 3D point clouds in the origin, and projecting them onto known planes.

In contrast to prior work that attempts to generalize into the real world by rendering real-world point clouds \cite{pointclipv2, pointcmt}, this method can be directly applied to real-world scene understanding, where the point cloud and images covering the $360^{\circ}$ view are available \cite{mei2022waymoopendatasetpanoramic}. The conditioning simplifies from ``conditioning with translation and orientation" to ``conditioning with orientation". As long as no foreign augmentation is applied to the original point cloud, the camera position will remain fixed, while images will be available from multiple directions. But now, the pitch parameter will be fixed. While there are more adaptations to be done, we ablate the learning ability on our existing datasets \cite{shapenet,objaverseXL} while using a single pitch value ($15^{\circ}$) and 6 yaw values in Table \ref{table:single-pitch}.

\begin{table}
\small
    \centering
        \caption{\textbf{Effect of fixing the pitch value.} Results show that fixing the pitch parameter at $15^{\circ}$ induces a slight drop in accuracy, but still produces a reliable representation model. In each case, the views are sampled randomly from the available sample space.}
    \resizebox{0.7\columnwidth}{!}{
    \begin{tabular}{lccc}
        \toprule
        \textbf{View Count} & \textbf{Fixed Pitch} & \textbf{Acc.} \\
        \midrule
        1 & \ding{55} & 93.76 \% \\
        1 & \ding{51} & 93.44 \% \\
        2 & \ding{51} & 93.35 \% \\
        3 & \ding{51} & 93.31 \% \\
        4 & \ding{51} & 93.31 \% \\
        5 & \ding{51} & 93.60 \% \\
        6 & \ding{55}  & 94.17 \% \\
        6 & \ding{51} & 93.27 \% \\
        12 & \ding{55} & 94.04 \% \\
        36 & \ding{55} & 93.71 \% \\
        \bottomrule
    \end{tabular}
    }
    \label{table:single-pitch}
\end{table}

Once the required image plane is conditioned at the predictor level, the conditioning can be further extended to predict patch-level representations of the image, making the learning objective more \emph{specific}. The availability of temporal point-cloud--image-set pairs \cite{mei2022waymoopendatasetpanoramic, Geiger2013IJRR} provides further avenues for conditioning.

It must also be noted that there is no \emph{conditioning limit} in image--point-cloud datasets. While predicting multiple views in parallel is limited by the available view counts, simply rotating the point cloud and camera angles (\emph{i.e., conditioning information}) from a continuous sampling space will be possible. This also opens up research for exploring the fitness of continuous positional encoding methods, such as Spherical Harmonics commonly seen in 3D reconstruction work \nocite{kerbl20233dgaussiansplattingrealtime}.

\end{document}